\RequirePackage{fix-cm}
\documentclass[twocolumn]{svjour3}          
\smartqed  
\usepackage{anyfontsize}
\usepackage{booktabs}
\usepackage{epsfig}
\usepackage{subfigure}
\usepackage{multirow}
\usepackage{array}
\usepackage{mathrsfs}
\usepackage{amsmath}
\usepackage{amssymb}
\usepackage{bm}
\usepackage{enumerate}
\usepackage{graphicx}
\usepackage[utf8]{inputenc}
\usepackage[english]{babel}
\usepackage[round, sort]{natbib}

\usepackage{setspace}
\usepackage{algorithm}
\usepackage{algpseudocode}
\usepackage{lipsum}
\usepackage{hyperref}
\usepackage{color}

\usepackage{makecell}
\usepackage{tabularx}
\usepackage{multirow}
\usepackage{pifont}
\usepackage{array}
\usepackage{subfigure}
\usepackage{bbding}

\begin{document}

\title{Vision Generalist Model: A Survey} 



\author{Ziyi~Wang \and
        Yongming~Rao \and
        Shuofeng~Sun \and
        Xinrun~Liu \and
        Yi~Wei \and
        Xumin~Yu \and
        Zuyan~Liu \and
        Yanbo~Wang \and
        Hongmin~Liu \and
        Jie~Zhou \and
        Jiwen~Lu$^\dagger$
}


\institute{ Ziyi Wang\textsuperscript{1}: wziyi22@mails.tsinghua.edu.cn \\
            Yongming Rao\textsuperscript{2}: raoyongming95@gmail.com \\
            Shuofeng Sun\textsuperscript{3}: ssf@bupt.edu.cn \\
            Xinrun Liu\textsuperscript{4}: d202110362@xs.ustb.edu.cn \\
            Yi Wei\textsuperscript{1}: y-wei19@mails.tsinghua.edu.cn \\
            Xumin Yu\textsuperscript{1}: yuxumin98@gmail.com \\
            Zuyan Liu\textsuperscript{1}: liuzuyan19@gmail.com \\
            Yanbo Wang\textsuperscript{1}: wyb23@mails.tsinghua.edu.cn \\
            Hongmin Liu\textsuperscript{4}: hmliu\_82@163.com \\
            Jie Zhou\textsuperscript{1}: jzhou@tsinghua.edu.cn \\
            Jiwen Lu\textsuperscript{1}: lujiwen@tsinghua.edu.cn \\ \\
            $^\dagger$ Corresponding Author\\
            \textsuperscript{1} Department of Automation, Tsinghua University, China \\
            \textsuperscript{2} Tencent HunyuanX, China \\
            \textsuperscript{3} Beijing University of Posts and Telecommunications, China \\
            \textsuperscript{4} University of Science and Technology Beijing, China
}

\date{Received: date / Accepted: date}

\maketitle

\begin{abstract}
Recently, we have witnessed the great success of the generalist model in natural language processing. The generalist model is a general framework trained with massive data and is able to process various downstream tasks simultaneously.  Encouraged by their impressive performance, an increasing number of researchers are venturing into the realm of applying these models to computer vision tasks. However, the inputs and outputs of vision tasks are more diverse, and it is difficult to summarize them as a unified representation. In this paper, we provide a comprehensive overview of the vision generalist models, delving into their characteristics and capabilities within the field. First, we review the background, including the datasets, tasks, and benchmarks. Then, we dig into the design of frameworks that have been proposed in existing research, while also introducing the techniques employed to enhance their performance. To better help the researchers comprehend the area, we take a brief excursion into related domains, shedding light on their interconnections and potential synergies. To conclude, we provide some real-world application scenarios, undertake a thorough examination of the persistent challenges, and offer insights into possible directions for future research endeavors.

\end{abstract}

\section{Introduction}

As an intelligence system, our human brain is able to perceive information from different input modalities and process multiple tasks simultaneously. Similar to human beings, in deep learning, the generalist model ~\cite{jaegle2021perceiverio, bae2022graphperceiverio,huang2023kosmos1, shukor2023unival} is a generic framework to handle various tasks without task-specific design. Recently, benefiting from the power of big data, large language models (LLMs) \cite{ouyang2022training,devlin2018bert,elmo} show the great success of the generalist models in natural language processing (NLP) areas. However, different from NLP, the output formats of vision tasks are diverse and complex. For example, the conventional classification methods \cite{russakovsky2015imagenet, he2016deep} only need to give the category of the image or point cloud. The object detection models further locate objects, and the outputs are bounding boxes. Segmentation models produce per-pixel semantic masks. Thus, for the vision generalist models \cite{hu2021UniT,zhang2023metatransformer,zhu2022uniperceiver}, it is essential to deploy a system that can be adapted to a wide range of vision downstream tasks.  

Compared to conventional neural networks, generalist models obtain billions of parameters and are trained with huge amounts of data, enjoying outstanding properties that are absent in conventional ones. Specifically, vision generalist models have the following advantages:

\textbf{1) Zero-shot multi-task transfer:} For different computer vision tasks, conventional methods often adopt task-specific frameworks to deal with various inputs and outputs. To process multiple tasks, multi-task learning methods \cite{zhang2021survey,sener2018multi,yu2020gradient} have been developed in the past few years. However, these methods fail to generalize to novel datasets without fine-tuning. Pretrained on huge amounts of task-agnostic data, the generalist models learn general representations and are able to be extended to diverse downstream tasks. Moreover, the generalist models have zero-shot transfer ability and do not need a specific adapter to perform fine-tuning, which achieves the goal of general perception.

\textbf{2) Multimodality inputs:} One of the properties of the generalist models is that they can receive multimodal data as inputs. Since the gap between different data modalities is huge, it is difficult to encode them as unified features. For example, images are regular 2D matrices while point clouds are unordered 3D vectors. The classical encoders of these two types of data are also different: 2D convolutions and 3D sparse convolutions \cite{graham20183d,yan2018second}. Except for vision signals, other modalities such as text and audio should also be considered, which further increases the difficulties.  Thanks to the Transformer architecture \cite{vaswani2017transformer}, some works unify the multimodality inputs as a series of tokens. 

\textbf{3) Great representation ability:} The existing generalist models often obtain billions of parameters. Although computationally expensive, the huge parameter number greatly improves models' representation ability. Multiple tasks and multimodal inputs can facilitate each other and further boost model capacity.  

\textbf{4) The power of big data:} Big data provides enormous knowledge to train the models. For instance, ChatGPT \cite{ouyang2022training} is trained with about 45TB of text data. Massive data collected from different modalities and domains increases the diversity of training samples, further prompting the generalization ability of generalist models. The large-scale datasets \cite{krizhevsky2012imagenet, chen2015coco} cover corner cases, helping the model to deal with extreme scenarios. 

\begin{figure*}[t]
    \begin{center}
    \includegraphics[width=\linewidth]{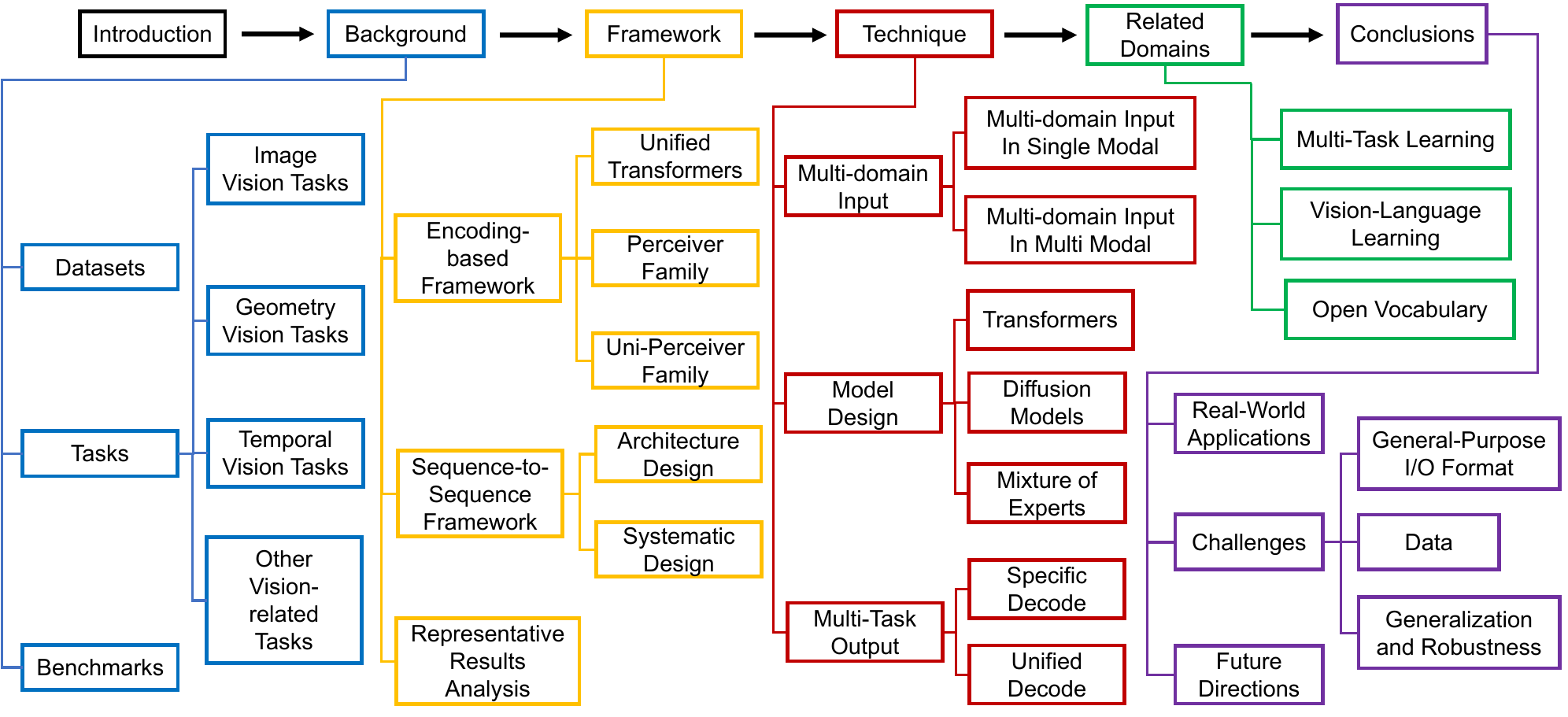}
    \vspace{-15pt}
    \caption{\textbf{The overview of the survey.} In Section~\ref{sec:background}, we first introduce the groundwork, including datasets, tasks, and benchmarks. Section~\ref{sec:framework} gives an analysis of popular framework design. The techniques involved in multi-domain inputs, model design, and multi-task outputs are shown in Section~\ref{sec:technique}. To better help the researchers understand the area, we further review the related domains in Section~\ref{sec:related-domains}. Section~\ref{sec:conclusions} provides applications, challenges, and future directions of vision generalist models.}
    \label{fig:overview}
    \end{center}
    \vspace{-15pt}
\end{figure*}

Although vision generalist models have many advantages, they face several challenges currently:

\textbf{1) Framework design:} A key technique of the generalist model is to design a general framework to process multiple downstream tasks. Several works \cite{hu2021UniT,zhang2023metatransformer,zhu2022uniperceiver} attempt to address this problem.  However, a standard pipeline does not exist in this field. Therefore, the most urgent challenge that needs to be solved is building a unified paradigm of the vision generalist model.

\textbf{2) Data acquisition:} Massive training data is necessary for the generalist models. In the NLP area, vast amounts of text data and corresponding annotations can be obtained from web pages. Unlike NLP, most CV data collected from the Internet does not have labels. It is extremely expensive and time-consuming to annotate this kind of data. Some existing works \cite{ouyang2022training,kirillov2023segment} give potential solutions to automatically label the data of specific tasks. However, how to automatically annotate the data from different tasks and modalities is still underexplored.

\textbf{3) Ethical risks:} Similar to LLM models, vision generalist models also have some ethical risks. For generation tasks, the models may produce content that contains personal or sensitive information, such as deepfake videos \cite{guera2018deepfake, westerlund2019emergence}. For discrimination tasks, the unintended bias in massive data will affect the models' decisions. Moreover, some questionable or dirty data may lead to legal issues. 

In the past two years, we have witnessed the success of generalist models in many deep learning topics. With the development of neural network architectures, more and more works have begun to focus on designing a general model to perform universal perception. Although the generalist model has attracted much research interest, there lack of a comprehensive review on this hot topic, which motivates us to write this survey. Specifically, the purposes of this paper can be summarized as follows: 1) Giving a literature review of this topic, which is convenient for the researchers to quickly get familiar with the area. 2) Figuring out the challenges and limitations of current methods and introducing future research directions. 3) Summarizing the differences and relations with other domains. 

Related to this paper, the survey \cite{awais2023foundational} gives an overview of vision foundation models. Similar to the vision generalist model, the foundation model is also trained with large-scale broad data, equipped with the capability to receive input signals from multiple modalities. However, the generalist models have a strong generalization ability to process multiple tasks simultaneously, which the foundation model lacks. When adapting to downstream tasks, foundation models need to be fine-tuned on specific datasets, restricting their applications in real-world scenarios. Therefore, our survey is conceptually different from~\cite{awais2023foundational} and we focus more on summarizing general modality perception and general task handling. In contrast to the broader perspective offered by another recent survey~\cite{li2023multimodal}, which comprehensively explores the taxonomy and evolution of multimodal foundation models, encompassing Unified Vision Models, Large Language Models, and their applications in multimodal agents, our survey maintains a narrower focus. We delve exclusively into the realm of vision generalist models, providing a detailed examination of the frameworks and techniques employed in their construction.

We structure this survey into six chapters to provide a systematic overview of the vision generalist model (VGM), shown in Figure \ref{fig:overview}. In Section~\ref{sec:background}, we lay the groundwork by introducing the general tasks, datasets, and benchmarks commonly employed in VGM research. Section~\ref{sec:framework} delves into the intricate framework designs that underpin VGM, classifying them into encoding-based methods and sequence-to-sequence frameworks. Moving forward, Section~\ref{sec:technique} takes a closer look at the various techniques devised to tackle the intricate challenges presented by multi-domain inputs, model design, and multi-task outputs. Section~\ref{sec:related-domains} broadens the scope by exploring the synergistic connections between VGM and related domains, specifically examining multi-task learning, vision-language learning, and open vocabulary learning. Finally, in Section~\ref{sec:conclusions}, we shed light on real-world applications and discuss the challenges faced by VGM, charting a course for future developments. We aim to offer a comprehensive resource for researchers and practitioners interested in the rapidly evolving landscape of vision generalist models.

\section{Background}
\label{sec:background}
The quantity and diversity of tasks (along with their associated datasets and benchmarks) that a vision generalist model can accommodate and process is a robust indicator of the model's generalization capacity. In this section, we provide an in-depth exposition of the background of the vision generalist model from three basic aspects: datasets, tasks, and benchmarks.

\subsection{Datasets}
Datasets used for pre-training or training vision generalist models are typically obtained through two primary approaches: utilizing existing vision datasets or creating large-scale datasets tailored to the design of the generalist architecture. This subsection focuses on the latter, as existing datasets are closely tied to specific vision tasks and will be comprehensively analyzed in Section~\ref{sec:tasks} and Table~\ref{tab:dataset}. For newly introduced large-scale datasets, we first discuss their data collection and organization processes, followed by methods employed for efficient annotation creation.

\vspace{4pt}
\noindent\textbf{LTIP, VTP, M3W.} Flamingo~\cite{alayrac2022flamingo} introduces three large-scale datasets for pre-training generalist vision models. LTIP (LongText and Image Pairs) contains 312 million text-image pairs, and VTP (Video and Text Pairs) has 27 million video-text pairs, both sourced from publicly available web data and cleaned with various software tools. M3W (MultiModalMassive-Web) includes 43.3 million instances, 185 million images, and 182 GB of text, following a similar collection process to the MassiveWeb~\cite{rae2021massiveweb} dataset. After filtering out non-English documents and those with explicit content, a custom scraper was used to extract relevant text content interleaved with images.

\vspace{4pt}
\noindent\textbf{WebLI.} PALI~\cite{chen2022pali} introduces WebLI, a multilingual image-text dataset that expands to 109 languages, covering 10 billion images and 12 billion text alternatives. The data is scraped from the public web, then heavily filtered and post-processed. To balance quality and scale, the dataset retains only the top 10\% of image-text pairs, resulting in about 1B examples. Additionally, OCR annotations are extracted for all images using a publicly available service, resulting in 29 billion image-OCR pairs.

\vspace{4pt}
\noindent\textbf{GRIT.} Kosmos-2~\cite{peng2023kosmos2} introduces GRIT (Grounded Image-Text pairs), which constructs image-text pairs by associating noun phrases and referential expressions in images with corresponding regions. The GRIT dataset contains approximately 91 million images, 115 million text phrases, and 137 million bounding boxes. The data collection process consists of three stages: (1) Generating noun phrase-bounding box pairs: noun phrases are extracted from image-text pairs and linked to image regions using a pre-trained detector. (2) Denoising: abstract noun phrases are excluded to reduce noise. (3) Bounding box selection: a pre-trained model is used to obtain bounding boxes, and non-maximum suppression removes highly overlapping boxes, retaining those with a confidence score above 0.65. The final annotations include noun phrases, referential expressions, and bounding boxes.

\vspace{4pt}
\noindent\textbf{UVDv1.} LVM~\cite{bai2024lvm} introduces UVDv1 (Unified Vision Dataset v1), which combines various types of visual data, including unlabeled images, images with visual annotations, unlabeled videos, videos with visual annotations, and 3D synthetic objects. UVDv1 contains 1.64 billion images, with over 80\% of these being unlabeled, providing a large dataset for model training and evaluation.

\vspace{4pt}
\noindent\textbf{MUSE.} 
PixelLM~\cite{ren2024pixellm} introduces MUSE (Multi-target Reasoning Segmentation), designed for complex pixel reasoning tasks, particularly multi-object segmentation. The MUSE dataset contains 910,000 high-quality instance segmentation masks with detailed text descriptions, sourced from the LVIS dataset for rich and diverse data. On average, each question-answer pair involves 3.7 objects, with a total of 246,000 pairs created. The dataset was generated by first using GPT-4 for image annotation, followed by the generation of questions about multiple image regions. To enhance data diversity, GPT-4V, a model adept at understanding visual content, was used. Annotations include instance-level mask annotations, text descriptions, and multi-objective question-answering results.

\begin{table*}[htbp]
	\centering
    \setlength\tabcolsep{0.5pt}
	\caption{Existing task-related datasets that are usually used by vision generalist models.}
        \fontsize{5.6}{10}\selectfont 
	\begin{tabular}{cc}
    \toprule
        \multicolumn{1}{c}{\footnotesize \centering Task} & \multicolumn{1} {c}{\footnotesize \centering Dataset} \\
	\midrule
        \multicolumn{2}{c}{\footnotesize \centering \emph{Image Vision Tasks}}\\
    \midrule
		\multicolumn{1}{c}{Image Classification } & ImageNet, CIFAR-10/100, Food101, DTD, Oxford-Pets, Flowers102, Place365, ObjectNet, ReaL, ILSVRC, CUB \\
        \multicolumn{1}{c}{Image Object Detection } & COCO, Object365, Visual Genome, LVIS, PASCAL VOC \\
        \multicolumn{1}{c}{Image Segmentation }  & COCO, PASCAL VOC, ADE20K, Hypersim, PhraseCut, Taskonomy, Pascal-5i \\
        \multicolumn{1}{c}{Keypoint Detection  } & Taskonomy, Pascal 3D+ \\
        \multicolumn{1}{c}{Visual Question Answering} & VQA, VQAv2, OKVQA, GQA, TallyQA, STVQA, VQA Karpathy split, VizWizVQA, VizWizG, xGQA, MaXM \\
        \multicolumn{1}{c}{Image Captioning} & COCO Caption, NoCaps, TextCaps, VizWiz-Cap, Screen2Words, Widget-Cap \\
        \multicolumn{1}{c}{Visual Commonsense Reasoning} & VCR \\
        \multicolumn{1}{c}{Image-Text Retrieval} & Flickr30k, COCO Caption \\
        \multicolumn{1}{c}{Referring Expression Comprehension} & RefCOCO, RefCOCO+, RefCOCOg, Visual7W, GuessWhat \\
        \multicolumn{1}{c}{Text-to-Image Generation} & COCO Caption \\
    \midrule
        \multicolumn{2}{c}{\footnotesize \centering \emph{Geometry Vision Tasks}}\\
    \midrule
        \multicolumn{1}{c}{Depth Estimation} & NYUv2 \\
        \multicolumn{1}{c}{Point Cloud Object Classification} & ModelNet40, ScanObjectNN \\
        \multicolumn{1}{c}{Point Cloud Part Segmentation}   & ShapeNetPart \\
        \multicolumn{1}{c}{Point Cloud Segmentation}   & SUN RGB-D, ScanNet, S3DIS, ShapeNet, NYU-v2 \\
    \midrule
        \multicolumn{2}{c}{\footnotesize \centering \emph{Temporal Vision Tasks}}\\
    \midrule
        \multicolumn{1}{c}{Optical Flow Prediction}   & Sintel, KITTI \\
        \multicolumn{1}{c}{Video Classification\& Recognition } & Kinetics-400, Kinetics-600, UCF101, Something-Something v2, EPIC-KITCHENS-100\\
        \multicolumn{1}{c}{Video Captioning}   & MSVD, YouCook2, VATEX, MSRVTT, ActivityNet Captions, Spoken Moments in Time \\
        \multicolumn{1}{c}{Video Question Answering}  & NExTQA, MSRVTTQA, ActivityNetQA, Social-IQ, iVQA, TGIF-Frame \\
        \multicolumn{1}{c}{Text-Video Retrieval}   & MSRVTT, MSVD, YouCook2, VATEX, DiDeMo, LSMDC \\
        \multicolumn{1}{c}{Text-to-Video Generation}   & CelebV-Text, CMU-MOSI \\
        \multicolumn{1}{c}{Egocentric Video Understanding} & Ego4D, EPIC-Kitchens, EGTEA Gaze+, Charades-Ego, THU-READ \\
        \multicolumn{1}{c}{Vision-Language Navigation} & R2R, R4R, VLN-CE, REVERIE, SOON, CVDN \\
	\bottomrule
	\end{tabular}%
	\label{tab:dataset}%
    \vspace{-5pt}
\end{table*}%

\subsection{Tasks}
\label{sec:tasks}
We systematically classify the vision tasks addressed by existing vision generalist models into four categories: image-based tasks, geometry-based tasks, temporal tasks, and other vision-related tasks. Although these models are also capable of handling non-vision tasks such as text classification or audio-related tasks like audio captioning, we exclude them from this discussion as our focus is primarily on vision data and tasks. For reference, task-related datasets are summarized in Table~\ref{tab:dataset}. 

\subsubsection{Image Vision Tasks}

\noindent\textbf{Vision-only Tasks.} Image vision tasks are broadly categorized into two main streams: perception and generation. Most vision generalist models primarily target perception tasks, with only a few recent models~\cite{lu2024unifiedio2, bai2022ofasys, wu2024visionllmv2} exploring capabilities in image generation. Perception tasks can be further classified by granularity: (1) Image-level perception, including tasks such as image classification; (2) Region-level perception, encompassing object detection, edge detection, crowded pedestrian detection, and human pose estimation; and (3) Pixel-level perception, which includes semantic segmentation, instance segmentation, and keypoint detection.

\vspace{4pt}
\noindent\textbf{Vision-Language Tasks.} When combined with text, vision-language tasks enable a wide range of applications in human-computer interaction. Some tasks require text outputs, such as visual question answering, image captioning, grounded captioning, visual commonsense reasoning, and descriptive image classification. Others focus on generating visual results, including image-text retrieval, visual grounding, referring expression comprehension, referring segmentation, text-to-image generation, and instruction-based image editing.

\subsubsection{Geometry Vision Tasks}

\noindent\textbf{Vision-only Tasks.} In 3D vision, vision-only tasks commonly addressed by vision generalist models can be categorized into two groups: (1) Low-level tasks, such as depth estimation and surface normal prediction; and (2) High-level tasks, including point cloud object classification, point cloud part segmentation, and point cloud semantic segmentation.

\vspace{4pt}
\noindent\textbf{Vision-Language Tasks.} Although the modality bridge between 3D vision and text remains underdeveloped, recent studies have made some progress~\cite{xu2024pointllm}. For instance, 3D visual question answering has been demonstrated, where point clouds are used as input to generate textual answers to 3D-related queries. Another notable application is robot manipulation~\cite{reed2022gato, driess2023palme, huang2023leo}, which requires integrating language understanding for interpreting instructions with a geometric understanding of the environment to execute actions.

\subsubsection{Temporal Vision Tasks}

\noindent\textbf{Vision-only Tasks.} Video tasks can be grouped into three main categories: (1) Motion understanding, encompassing tasks like optical flow prediction and action recognition; (2) Semantic perception, including video classification; and (3) Temporal prediction, such as video frame prediction. While these tasks primarily focus on conventional third-person videos, such as those in datasets like Kinetics, egocentric video understanding has attracted increasing attention. This line of research centers on first-person video analysis to model human-object interactions, gaze behavior, or temporally grounded activities, making it particularly relevant for embodied AI and real-world human-centric applications.

\vspace{4pt}
\noindent\textbf{Vision-Language Tasks.} Similar to image vision tasks, video perception, and generation can also benefit from language integration. Tasks such as video captioning and video question answering produce textual outputs, while text-video retrieval and text-to-video generation focus on creating visual results. In addition, vision-language navigation has emerged as a temporal and embodied task, where agents interpret natural language instructions to navigate 3D environments. It formulates a systematic challenge that integrates visual perception, language grounding, and sequential decision-making.

\subsubsection{Other Vision-related Tasks}
Beyond the vision tasks for 2D images, 3D geometry, and temporal videos discussed earlier, there are various other tasks that utilize visual data for broader applications. For instance, Kosmos-1~\cite{huang2023kosmos1} and LVM~\cite{bai2024lvm} demonstrate generalized intelligence by performing human IQ tests. Similarly, Perceiver-IO~\cite{jaegle2021perceiverio} showcases its versatility on discrete modalities by solving game-related and symbolic representation tasks in StarCraft II.

\subsection{Benchmarks and Evaluation}
While previous evaluations of vision generalist models often focused on individual tasks and datasets, recent research has shifted towards benchmarks that incorporate multiple datasets to more accurately assess the generalization capabilities of these models. We believe that composite evaluation, which evaluates models across various datasets simultaneously, provides a more comprehensive comparison than reporting performance on multiple datasets separately. Here, we briefly introduce some representative comprehensive benchmarks.

\vspace{4pt}
\noindent\textbf{MMMU.}
The MMMU benchmark (Multimodal Multitask Understanding)~\cite{yue2024mmmu} is a large-scale evaluation suite aimed at assessing the reasoning and comprehension capabilities of multimodal models on college-level tasks. It consists of 11.5K multimodal questions derived from authentic academic sources such as university exams, course materials, and textbooks. The benchmark spans six broad domains: Art \& Design, Business, Science, Health \& Medicine, Humanities \& Social Science, and Tech \& Engineering, covering 30 subject areas and 183 specialized subfields. The dataset incorporates 30 distinct image formats, including visualizations like diagrams, maps, charts, tables, musical notation, and molecular structures.

\vspace{4pt}
\noindent\textbf{GRIT.} GRIT (General Robust Image Task)~\cite{gupta2022gritbm} assesses the performance, robustness, and calibration of vision systems across various image prediction tasks, concepts, and data sources. The benchmark includes seven tasks that evaluate a wide range of vision skills: object classification, object localization, reference representation grounding, visual question answering, segmentation, human keypoint detection, and surface normal estimation.

\vspace{4pt}
\noindent\textbf{VIMA-Bench.}
VIMA-Bench~\cite{jiang2022vima} is a benchmark suite for multimodal robot learning, designed to provide diverse tasks and a comprehensive testbed for evaluating agent capabilities. It supports various task specifications, including imitation learning, language instructions, and visual goals. By extending the Ravens robot simulator, VIMA-Bench generates numerous task instances, covering 17 meta-tasks and thousands of specific tasks.

\vspace{4pt}
\noindent\textbf{MMBench.}
MMBench~\cite{liu2025mmbench} is a multimodal benchmark to evaluate large visual-language models (VLMs), addressing the limitations of traditional benchmarks in evaluation depth and accuracy. It includes over 3,000 multiple-choice questions spanning 20 capability dimensions, such as object localization and social reasoning. Each dimension contains more than 125 questions, ensuring a comprehensive and balanced evaluation. MMBench-Video\cite{fang2024mmbenchvideo} is introduced to assess the video understanding capabilities of VLMs, incorporating approximately 600 long-form videos sourced from the web and 2,000 corresponding question-answer pairs. To mitigate evaluation bias, it leverages GPT-4\cite{achiam2023gpt4} to focus more on semantic consistency rather than surface-level phrasing differences.

\vspace{4pt}
\noindent\textbf{Long-video Benchmarks.}
In addition to MMBench-Video, which exclusively targets multimodal question answering, several recent benchmarks have been proposed to evaluate long-video understanding from different perspectives. LongVideoBench\cite{wu2024longvideobench} introduces a referring reasoning task to assess the long-context multimodal reasoning capabilities of VLMs, where specific video segments and referring queries are provided. The benchmark comprises 6,678 multiple-choice questions across 3,763 videos, some extending up to an hour in length. LVBench\cite{wang2024lvbench} emphasizes language grounding tasks, including captioning with temporal grounding and key information retrieval, using 500 videos each exceeding 30 minutes in duration. EgoSchema~\cite{mangalam2023egoschema} focuses on schema induction from egocentric videos, with clips of approximately 100 seconds in certificate length. Certificate length is a novel metric introduced in the benchmark, defined as the minimum video duration required for a human verifier to be confident in the correctness of the annotated event. This setup encourages the development of models capable of abstract schema learning and long-horizon temporal reasoning.

\vspace{4pt}
\noindent\textbf{MME.}
MME~\cite{fu2023mme} is a comprehensive benchmark for fair performance comparison among multimodal large language models (MLLMs), covering both perceptual and cognitive abilities across 14 subtasks. These include image recognition tasks (e.g., identifying the presence, quantity, location, and color of objects) and cognitive tasks such as common sense reasoning, numerical computation, text translation, and code reasoning. Recently, Video-MME~\cite{fu2024videomme} was introduced as the first comprehensive evaluation benchmark for MLLMs in video analysis, while MME-RealWorld~\cite{zhang2024mmerealworld} focuses on high-resolution real-world scenarios that are challenging even for humans. Additionally, the MME-Survey~\cite{fu2024mmesurvey} provides a thorough analysis of various multimodal LLM evaluations.

\vspace{4pt}
\noindent\textbf{MM-vet.}
MM-Vet~\cite{yu2023mmvet} is a benchmark aimed at systematically grading the performance of large multimodal models (LMMs) on complex vision-language tasks. It organizes evaluation around six foundational capabilities: visual recognition, optical character recognition, factual knowledge, language generation, spatial reasoning, and mathematical understanding. These capabilities are operationalized into 16 task types that collectively assess how well models can integrate and reason across multiple modalities.

\vspace{4pt}
\noindent\textbf{SEED-Bench.}
SEED-Bench~\cite{li2023seedbench} is a benchmark that quantitatively presents generative comprehension capabilities of multimodal large language models. It includes 19,000 manually annotated multiple-choice questions across 12 evaluation dimensions, including image and video comprehension. 

\vspace{4pt}
\noindent\textbf{Raven IQ.}
The Raven IQ test dataset~\cite{raven2003raven, carpenter1990raveniq} consists of 50 examples, each with a varying number of images, typically 3 (2×2 matrix), 4, or 8 (3×3 matrix). Each instance presents six candidate images, with only one being the correct completion. The model's task is to infer the next image based on the given ones. This test serves as an effective platform for evaluating a model's non-linguistic reasoning abilities and is based on Raven's Progressive Matrices, a widely used test for assessing intelligence quotient (IQ).

\section{Framework}
\label{sec:framework}

In this section, we conduct an in-depth survey of existing frameworks that have been put forth for vision generalist models. We heuristically categorize them into two perspectives: the encoding-based methods and the sequence-to-sequence frameworks. This systematic categorization aims to provide a comprehensive understanding of the current landscape of research in this domain, shedding light on the diverse strategies pursued to address the challenges of building a vision generalist model. We denote $\mathcal{E}_{X}$ and $\mathcal{D}_{X}$ as domain-specific encoders and task-specific decoders, respectively, where the arbitrary modality placeholder $X$ can be substituted with $V$, $L$, $A$, or $M$, representing vision, language, audio, or other custom modalities. Additionally, $I$, $O$, $F$, and $T$ are used to represent input data, output results, intermediate features, and encoded tokens, while $\mathcal{T}_E$ and $\mathcal{T}_D$ indicate Transformer encoder and decoder.

\subsection{Encoding-based Framework}

Encoding-based vision generalist models are primarily focused on devising a unified encoding space that can effectively accommodate different modalities and handle a wide array of tasks. Within this domain, three prominent frameworks have been identified, each pursuing unifying models with distinct approaches. One line of framework~\cite{hu2021UniT, zhang2023metatransformer, bachmann2022multimae}  leverages the concept of unified Transformers~\cite{vaswani2017transformer} as shared encoder-decoder models. These methods incorporate \\ domain-specific token encoders and task-specific output heads to address the multimodality and multi-task challenges effectively. Another line of framework~\cite{jaegle2021perceiverio, bae2022graphperceiverio, carreira2022hip} extends the concept of general perception by employing iterative attention mechanisms inspired by the Perceiver architecture~\cite{jaegle2021perceiver}. Uni-Perceiver family~\cite{zhu2022uniperceiver, zhu2022uniperceivermoe, li2023uniperceiverv2} explores the maximum likelihood between input modalities and target domains within a shared latent space.

\subsubsection{Unified Transformers}

\noindent \textbf{Preliminary.} Transformers~\cite{vaswani2017transformer} have revolutionized natural language processing and various other tasks involving sequential data. Transformers are built on the self-attention mechanism, enabling them to process inputs simultaneously in parallel rather than sequentially:
\begin{equation}
    \text{Attention}(Q, K, V) = \text{softmax}\left(\frac{QK^T}{\sqrt{d_k}}\right) V
\label{eq:attn}
\end{equation}
where $Q, K, V$ are the query, key, and value matrices, respectively. $\text{softmax}$ denotes the softmax function, and $d_k$ represents the dimensionality of the key vectors. This unique attention mechanism allows Transformers to capture long-range dependencies and contextual relationships effectively, making them highly efficient in handling large input sequences. The following works extend self-attention to cross-attention and build the Transformer decoder to handle instructed feature decoding with task-specific queries. 

\vspace{4pt}
\noindent \textbf{Unified Transformers as VGM.} 
Unified Transformers models initially design domain-specific encoders to convert multi-domain data into tokens, which are then aggregated through Transformer attention layers. Task-specific decoders are subsequently employed to handle multiple tasks concurrently:
\begin{equation}
    O_{X} = \mathcal{D}_{X}(\mathcal{T}_D(\mathcal{T}_E(\mathcal{E}_{X}(I_{X}))))
\end{equation}
Although these models address multiple domains and tasks, they are jointly trained on a shared Transformer architecture, promoting mutual learning across modalities and tasks.

Taking inspiration from the remarkable achievements of Transformers in specific domains, UniT~\cite{hu2021UniT} pioneers an investigation into the potential of Transformers to extract cross-domain knowledge and efficiently handle diverse tasks using a unified model. The framework utilizes a Transformer encoder architecture for both image and text encoders, differing only in their input projection layers and model weights. Subsequently, the encoded sequences from both domains are concatenated to form a unified memory bank within the domain-agnostic Transformer decoder. The query input for the UniT decoder consists of task-specific embeddings, and the output of the UniT decoder is further directed via task-specific heads, catering to the various tasks at hand. The model undergoes joint training on multiple datasets and tasks. 

While UniT focuses exclusively on image and text domains, MetaTransformer~\cite{zhang2023metatransformer} significantly broadens this concept by extending it to encompass up to 12 different modalities. To achieve this, they introduce a formal meta-tokenization scheme that encodes input data into token embeddings residing in a shared manifold space. During the modality-agnostic learning phase, the pre-trained unified Transformer encoder remains frozen, while the learnable classification token is continually updated to serve as the summary representation of each domain.

Off-the-shelf pre-training schemes designed for Transformers can also be effectively harnessed to enhance the representation capabilities of unified Transformer models. MultiMAE~\cite{bachmann2022multimae} pioneers this approach by pseudo-labeling various tasks on ImageNet-1K to create a multi-task dataset. During the pre-training phase, domain-specific Transformer decoders are introduced following the unified Transformer encoder to predict masked tokens. In the fine-tuning stage, task-specific decoders take over to predict task outputs, thereby tailoring the model's learning to the specific objectives of each task.

GenLV~\cite{chen2024genlv} introduces a visual task prompting framework to improve generalization across low-level vision tasks using pre-trained visual models. Tasks are specified via paired prompt images (source and target), which encode the desired transformation. These prompt representations are then fused with mid-level features from the foundation model through a novel prompt cross-attention mechanism.

In summary, the Transformer framework holds significant potential to function as a unified model and effectively extract cross-domain knowledge when data from diverse domains is encoded within a shared manifold. This capacity enables the model to address a wide range of tasks with task-specific heads, making it a promising avenue for building generalist vision models.

\begin{figure*}[t]
    \begin{center}
    \includegraphics[width=\linewidth]{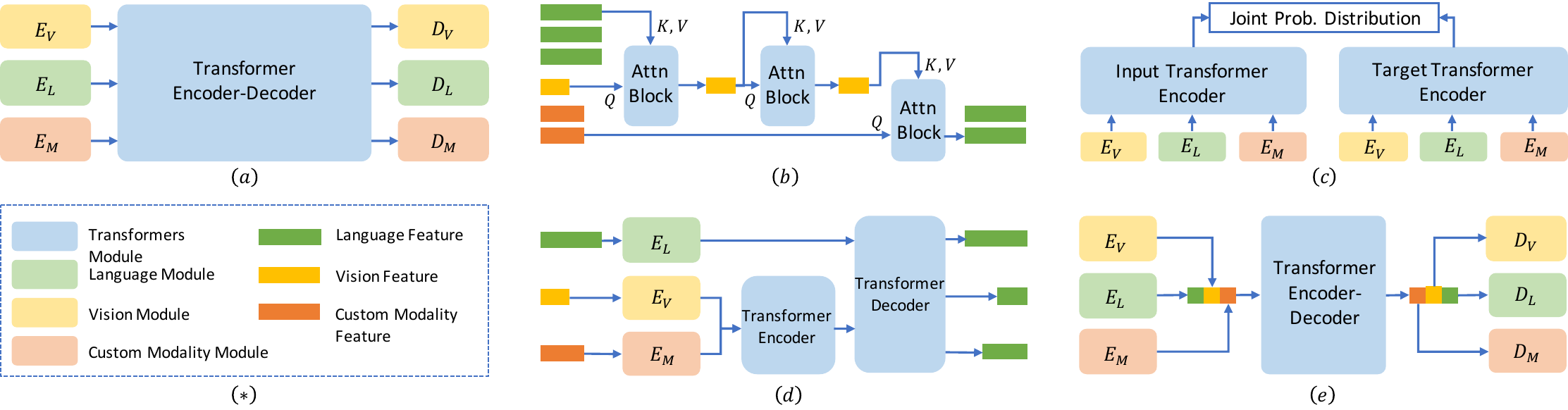}
    \vspace{-15pt}
    \caption{\textbf{Frameworks for generalist vision models.} Figure (a) shows unified Transformers as the generalist vision framework. Figure (b) shows the key principle of the Perceiver~\cite{jaegle2021perceiver} family. Figure (c) shows a general framework for the Uni-Perceiver~\cite{zhu2022uniperceiver} family. Figure (d) shows a language-modeling framework that unifies various tasks as the text generation task. Figure (e) shows a more generalized sequence-to-sequence architecture that can handle more modalities besides language, such as vision.}\label{fig:framework}
    \end{center}
    \vspace{-10pt}
\end{figure*}

\subsubsection{Perceiver Family}

\noindent \textbf{Preliminary.} Perceiver~\cite{jaegle2021perceiver} is a powerful and versatile model that achieves general perception capabilities across different modalities. It combines a cross-attention mechanism and an iterative refinement process. The cross-attention forms a bottleneck, enabling the model to process input sequences of varying lengths and modalities effectively. The refinement process improves the model's understanding through multiple iterations of attention and feedback loops. 

\vspace{4pt}
\noindent \textbf{Extending Perceiver as VGM.} 
While the Perceiver primarily concentrates on unifying input modalities and is limited to handling simple output spaces like classification, Perceiver-IO~\cite{jaegle2021perceiverio} takes a significant step forward by extending the concept of general perception to unify the output task space. This framework encodes multimodal data into features $F_{V}$, $F_{L}$, and $F_{X}$. Cross-attention is first applied between vision and language features, followed by self-attention within the vision modality for aggregation. When custom modalities are included, cross-attention is recursively applied to enable cross-modal interaction:
\begin{equation}
\begin{split}
    F_X &= \mathcal{E}_X(I_X) \\
    F^{\prime}_V &= \mathrm{Attention}(F_V, F_L, F_L) \\
    F^{\prime\prime}_V &= \mathrm{Attention}(F^{\prime}_V, F^{\prime}_V, F^{\prime}_V) \\
    O_X &= \mathcal{D}_X(\mathrm{Attention}(F_X, F^{\prime\prime}_V, F^{\prime\prime}_V))
\end{split}
\end{equation}
where the Attention operation follows Equation~\ref{eq:attn}, where $F^{\prime}$ and $F^{\prime\prime}$ represent intermediate features.

Based on the above formula, Perceiver-IO constructs queries using output-specific features and employs a cross-attention mechanism to generate outputs with distinct semantics tailored to different tasks. This advancement allows Perceiver-IO to address a broader range of tasks and further enhances its applicability in various domains. Its subsequent model, Graph Perceiver IO~\cite{bae2022graphperceiverio}, makes a progression in constructing the input array based on graph node features and the adjacency matrix that encapsulates topological information, further enhancing its capabilities in handling graph-related tasks. Another notable successor of the Perceiver model is Hierarchical Perceiver (HiP)~\cite{carreira2022hip}, which introduces the concept of locality by partitioning the input into blocks and developing hierarchical structures. In addition, HiP incorporates random masked auto-encoding (MAE) to facilitate the learning of low-dimensional positional embeddings, especially beneficial for handling high-resolution signals.

In conclusion, the Perceiver-style framework demonstrates the ability to efficiently compress input sequences of arbitrary length into fixed-length latent features, ensuring effective processing. Furthermore, it can accurately decode these latent features into sequences of varying lengths, accommodating the specific requirements of different tasks. This flexibility in handling variable-length inputs and outputs makes the Perceiver-style approach a powerful and adaptable solution for various applications across domains.

\subsubsection{Uni-Perceiver Family}

\noindent \textbf{Uni-Perceiver.} Despite sharing a similar name, the core concept of Uni-Perceiver~\cite{zhu2022uniperceiver} is distinct from the Perceiver model discussed previously. Uni-Perceiver presents a novel and innovative approach to achieving a generic perception architecture by building a shared representation space to unify mutlimodal inputs and targets. To achieve this, the framework incorporates lightweight modality-specific tokenizers to encode both inputs and targets and utilizes a modality-agnostic Transformer encoder to analyze the joint probability distributions between input and target sequential tokens. Consequently, different perception tasks are formulated in a unified manner, seeking to maximize the similarity between representations of inputs and targets. Specifically, the Uni-Perceiver family framework can be formulated as: 
\begin{equation}
    \begin{split}
    &\mathcal{L} = \sum_{i=1}^n \mathbb{E} \Bigg(-\mathrm{log}\frac{P(F_{X_i}, G_{X_i})}{\sum_{i\neq j}P(F_{X_i}, G_{X_j})}\Bigg) \\
    &P(f,g) \propto \mathrm{exp}(\mathrm{cos}(\mathcal{T}_E(f), \mathcal{T}_E(g))/\tau)
    \end{split}
\end{equation}
where $G$ represents task-specific features, $\tau$ is a hyperparameter, and $P(f,g)$ measures the similarity between multi-domain data and task representations.

\vspace{4pt}
\noindent \textbf{Successors of Uni-Perceiver.} 
To tackle the challenge of task interference in generalist models, Uni-Perceiver-MoE~\cite{zhu2022uniperceivermoe} presents a significant enhancement to the Uni-Perceiver model by introducing Conditional Mixture-of-Experts (MoE). This extension involves the incorporation of a gate decision vector, which effectively dispatches different input tokens to distinct experts based on task-specific conditions. Additionally, the paper explores various routing strategies to optimize the allocation of input tokens to the appropriate experts, addressing specific requirements of different tasks.

Uni-Perceiver v2~\cite{li2023uniperceiverv2}, on the other hand, places its emphasis on large-scale multi-task learning. To ensure stable multi-task learning, the authors propose Task-Balanced Gradient Normalization to address the challenge of gradient instability. They additionally introduce an unmixed sampling strategy to enable training with large batch sizes and facilitate efficient utilization of computational resources, making it well-suited for large-scale training scenarios involving multiple tasks from diverse domains.

In conclusion, the Uni-Perceiver family represents a significant advancement in generic perception frameworks, allowing for seamless integration and processing of diverse data modalities and multiple large-scale tasks within a unified framework.

\subsection{Sequence-to-Sequence Framework}

The concept of generalist models initially emerged in the domain of natural language processing, and the NLP community has developed several mature mechanisms for unifying cross-domain inputs and handling multiple types of tasks. Drawing inspiration from these mechanisms, various papers in the computer vision domain have proposed similar sequence-to-sequence frameworks for the design of generalist vision models. The architecture design strategies can be divided into prefix language modeling~\cite{cho2021vlt5, wang2021simvlm, wang2022omnivl, huang2023kosmos1, shukor2023unival, peng2023kosmos2, chen2023minigptv2,  wang2024visionllm}, masked language modeling~\cite{li2023lavender} and more generalized sequence generation~\cite{wang2022ofa, lu2022unifiedio, lu2024unifiedio2, ren2024pixellm, wu2024visionllmv2}. Besides, researchers have thoroughly investigated crucial systematic designing components of generic vision models, including how to scale up the model~\cite{chen2022pali, chen2023palix, bai2024lvm}, how to better migrate the pre-trained generic model to few-shot learning~\cite{alayrac2022flamingo}, how to design a sophisticated system for more user-friendly human-computer interaction~\cite{bai2022ofasys, liu2023interngpt}, and how to support embodied AI development~\cite{reed2022gato, driess2023palme, mu2024embodiedgpt, brohan2023rt, huang2023leo}.

\subsubsection{Architecture Design}

\noindent \textbf{Prefix Language Modeling.} PrefixLM is a conditional text generation approach that allows for bidirectional attention on the prefix sequence, while autoregressive factorization is applied only to the remaining tokens. This process can be formulated as:
\begin{equation}
    \begin{split}
        T_X &= \mathcal{E}_X(I_X) \\
        O_L &= \mathcal{T}_D(\mathrm{cat}(\mathcal{T}_E(T_X), T_L))
    \end{split}
\end{equation}where cat refers to concatenation, stacking modality-specific tokens as prefixes to language token sequences.

Inspired by T5~\cite{raffel2020t5}, VL-T5~\cite{cho2021vlt5} proposes to unify vision-language tasks via multimodal conditional text generation. VL-T5 implements a bi-directional multimodal encoder to fuse task instruction tokens and visual embeddings, whose output acts as a prefix to the autoregressive text decoder to generate text outputs. SimVLM~\cite{wang2021simvlm} also leverages the PrefixLM strategy to achieve task unification. In the case of image-text inputs, SimVLM utilizes image tokens as the prefix to predict text tokens, facilitating effective interaction between multiple modalities. Inspired by Florence~\cite{yuan2021florence}, OmniVL~\cite{wang2022omnivl} incorporates unified text generation tasks with unified contrastive learning to enhance the representation capacity of cross-modal features. A more recent generic model, Kosmos-1~\cite{huang2023kosmos1} also leverages the next-token prediction task as the training objective, with much larger web-scale multimodal corpora as training data and more advanced multimodal large language models to achieve better results. Its successor, Kosmos-2~\cite{peng2023kosmos2}, further enables new capabilities of perceiving object descriptions and grounding text to the visual world and integrates the grounding capability into downstream applications. UnIVAL~\cite{shukor2023unival} enhances the conditional next token prediction training by introducing multimodal curriculum learning and multimodal task balancing as the pre-training procedure,  obtaining a unified representation of various tasks. VisionLLM~\cite{wang2024visionllm} conceptualizes images as a foreign language, introducing a unified language-based instruction paradigm to address vision-centric tasks. The framework employs a language-conditioned image tokenizer alongside a task decoder built upon a large language model (LLM), enabling it to address open-ended tasks directed by natural language instructions. MiniGPTv2~\cite{chen2023minigptv2} proposes a unified interface for diverse vision-language tasks by introducing distinctive task identifiers to improve task differentiation. Additionally, it integrates spatial location representations, encoding spatial coordinates into a textual format for enhanced processing. Chameleon~\cite{team2024chameleon} unifies multiple modalities as discrete tokens within a mixed modal auto-regressive LM architecture. For mixed-modal generation, it employs an image de-tokenizer to reconstruct visual outputs. The model incorporates architectural enhancements and training strategies to stabilize the scalable training of early-fusion token-based models.

\vspace{4pt}
\noindent \textbf{Masked Language Modeling.} LAVENDER~\cite{li2023lavender} presents a comprehensive exploration of a unified video-language model, wherein MLM serves as the unifying interface for diverse tasks. For input unification, LAVENDER introduces a multimodal fusion encoder to compute cross-modal representations in addition to modality-specific encoders. The pre-training stage encompasses MLM and Video Text Matching tasks, both designed to predict the mask token in texts. In downstream adaptation, LAVENDER inserts or replaces existing tokens with mask tokens for consistent cross-entropy supervision of downstream tasks.

\vspace{4pt}
\noindent \textbf{Generalized Sequence Generation.} While language modeling mainly focuses on text sequences, more generalized sequence generation methods regard any data format or tasks of any modality as sequences. Their architecture mainly consists of two main components: an encoder and a decoder. The encoder ingests the input sequence and transforms it into a fixed-length sequence of tokens, commonly referred to as a context vector. This context vector is then utilized by the decoder, which generates the output sequence in an autoregressive manner, producing one token at a time:
\begin{equation}
    \begin{split}
        & T_X = \mathcal{E}_X(I_X) \\
        & S_I = \mathrm{cat}_i(T_{X_i}) \\
        & S_O = \mathcal{T}_D(\mathcal{T}_E(S_I)) \\
        & O_X = \mathcal{D}_X(\mathrm{split}_X(S_O))
    \end{split}
\end{equation}
where $S_I$ and $S_O$ denote input and output sequences, respectively. The cat operation concatenates tokens from different modalities into a unified sequence, while the split operation divides $S_O$ into modality-specific token sequences. Seq2Seq models are well-suited for handling variable-length input and output sequences. Their adaptability and strength in modeling distant dependencies have earned broad recognition.

As implied by its name, OFA (One For All)~\cite{wang2022ofa} devises a modality-agnostic and task-agnostic framework that accommodates a wide range of tasks. It establishes a general sequence-to-sequence architecture and adopts instruction-based learning in both the pre-training and fine-tuning stages, thereby eliminating the need for additional task-specific layers in downstream tasks. In particular, OFA employs patch sequential tokens to encode input images and utilizes byte-pair encoding for processing linguistic inputs. Target images are discretized via image quantization, while target objects are represented as location tokens of bounding boxes in addition to label tokens. Deliberately crafted instructions guide the learning process, encompassing multiple cross-domain tasks within the framework. 

Unified-IO~\cite{lu2022unifiedio} further extends the sequence-to-sequence framework to encompass a broader range of tasks and concurrently conducts joint training on a more extensive set of multi-domain datasets. To better encode discrete representations of images, the model adopts a pre-trained VQ-GAN~\cite{esser2021vqgan}, enhancing the representation ability and effectiveness of the framework. The successor, Unified-IO v2~\cite{lu2024unifiedio2}, expands the modality scope to encode text, images, audio, video, and interleaved sequences while generating outputs such as text, actions, audio, images, and sparse or dense labels. Initially built from scratch with varied multimodal data, the model is then fine-tuned via instruction-driven learning on a large corpus.

Building on VisionLLM~\cite{wang2024visionllm}, VisionLLMv2~\cite{wu2024visionllmv2} significantly broadens its applicability to encompass hundreds of vision-language tasks, overcoming the limitations of visual question answering (VQA). This advancement is enabled by a newly introduced information transmission mechanism called "super link," which employs Routing Tokens to define vision tasks and Super-Link Queries to extract task-specific information.

Departing from the sequential modeling approach commonly used in previous research, PixelLM~\cite{ren2024pixellm} emphasizes enhancing LLMs for pixel-level visual perception tasks. It introduces a lightweight pixel decoder and a comprehensive segmentation codebook as key innovations. The segmentation codebook, central to the framework, encodes pixel-level knowledge across various visual scales, thereby eliminating the reliance on external segmentation models.

Another recent effort to extend auto-regressive decoding in LLMs to diverse visual tasks is GiT~\cite{wang2024git}. It introduces a unified language interface that represents visual targets (such as segmentation masks and bounding boxes) as token sequences. GiT also proposes a flexible multi-task prompting template to unify task formats. The entire multi-layer Transformer is trained from scratch without relying on task-specific heads.

\subsubsection{Systematic Design}
\label{sec:system_design}

\noindent \textbf{Scalability Discussion.} 
PALI~\cite{chen2022pali} and PALI-X~\cite{chen2023palix} focus on examining the scalability of generic vision models trained through the unified text generation task. PALI emphasizes the crucial nature of jointly scaling both the vision and language components. They develop a sizable multilingual mix comprising eight pre-training tasks, built upon their newly introduced WebLI dataset, which encompasses 10 billion images and texts in over 100 languages. Subsequently, PALI-X further scales up both the vision and language components of PALI and observes a substantial performance improvement across a diverse array of tasks. By incorporating a blend of prefix completion and masked-token prediction tasks, the revised training strategy enhances overall scalability and efficacy. LVM~\cite{bai2024lvm} advances the scalability of generalized vision models by structuring raw visual data and annotations into "visual sentences," leading to the creation of the Unified Vision Dataset v1 (UVDv1). These visual sentences are processed through a large visual tokenizer and analyzed using an autoregressive Transformer model. This unified formatting approach enables LVM to fully harness the potential and significance of unsupervised visual data.

\vspace{4pt}
\noindent \textbf{Prompt Tuning for Few-shot.} Flamingo~\cite{alayrac2022flamingo} is dedicated to addressing the challenges of adapting generic models, trained through text generation, to downstream tasks under the few-shot setting. They introduce the concept of text-based prompt tuning, wherein few-shot text-image examples are treated as the prefix condition for autoregressive generation. The paper proposes novel perceiver resamplers and gated attention layers as prompt tuning parameters. These parameters enable effective adaptation of pre-trained large language models with frozen weights to various downstream applications, facilitating their versatility and performance under the few-shot learning paradigm.

\vspace{4pt}
\noindent \textbf{System.} OFASys~\cite{bai2022ofasys} introduces a streamlined and versatile user interface termed multimodal instruction. It presents a modularized and reusable system design to streamline the research process for multimodal multi-task learning. By employing the instruction interface, users can effortlessly declare a new task with a single line of code or easily customize task-specific processing and incorporate new modalities. This user-friendly design empowers researchers to efficiently experiment with multimodal multi-task learning, simplifying the integration of new tasks and modalities into the framework. OFASys shows a significant stride in the systematic exploration of generalist vision models. InternGPT~\cite{liu2023interngpt} supports chat-based interaction enhanced with non-verbal signals like pointing, facilitating direct user control over images and videos displayed on screen. The system comprises three main components: a perception module, such as SAM~\cite{kirillov2023sam}, to process user pointing instructions; an LLM controller, such as LLaMa~\cite{touvron2023llama} or GPT-4~\cite{achiam2023gpt4}, for language-based interaction; and an open-world toolkit, including options like ControlNet~\cite{zhang2023controlnet}, or InternVideo~\cite{wang2022internvideo}, to execute multimodal tasks effectively.

\vspace{4pt}
\noindent\textbf{Embodied AI.} In addition to the vision systems discussed earlier, some research focuses on the development of embodied AI systems through sequence-to-sequence vision-generalist models. GATO~\cite{reed2022gato} introduces the first multimodal, multi-task, and multi-embodiment generalist policy. This model encodes input data from multiple sources into tokens and trains an autoregressive Transformer model, using prepended task-specific prompts to define embodied tasks and derive corresponding actions. During deployment, the model leverages historical observations to predict future actions, which are then sent to the environment, resulting in new observations. PaLM-E~\cite{driess2023palme} builds upon this idea by utilizing a pre-trained large language model (LLM) to perceive multimodal input and predict sequential decisions in natural language, which are subsequently interpreted as actions by an embodied agent. RT-2~\cite{brohan2023rt} introduces the concept of vision-language-action (VLA) models, where actions are treated as text tokens and directly integrated with language and image tokens, allowing for a unified model to handle both perception and action prediction. EmbodiedGPT~\cite{mu2024embodiedgpt} proposes a large-scale embodied planning dataset, EgoCOT, and applies prefix tuning on pre-trained LLMs. The model also extracts task-relevant features from LLMs to facilitate both high-level planning and low-level control, enhancing the agent's ability to perform complex tasks. LEO~\cite{huang2023leo} further extends the application of vision-generalist models to embodied agents in the 3D domain. Their approach first ensures a robust 3D vision-language alignment and then applies 3D vision-language-action instruction tuning. They also introduce two novel datasets, LEO-align and LEO-instruct, corresponding to these two distinct stages of training, respectively, advancing the capabilities of embodied agents in spatially complex environments.

\begin{figure}
    \includegraphics[width=0.5\textwidth]{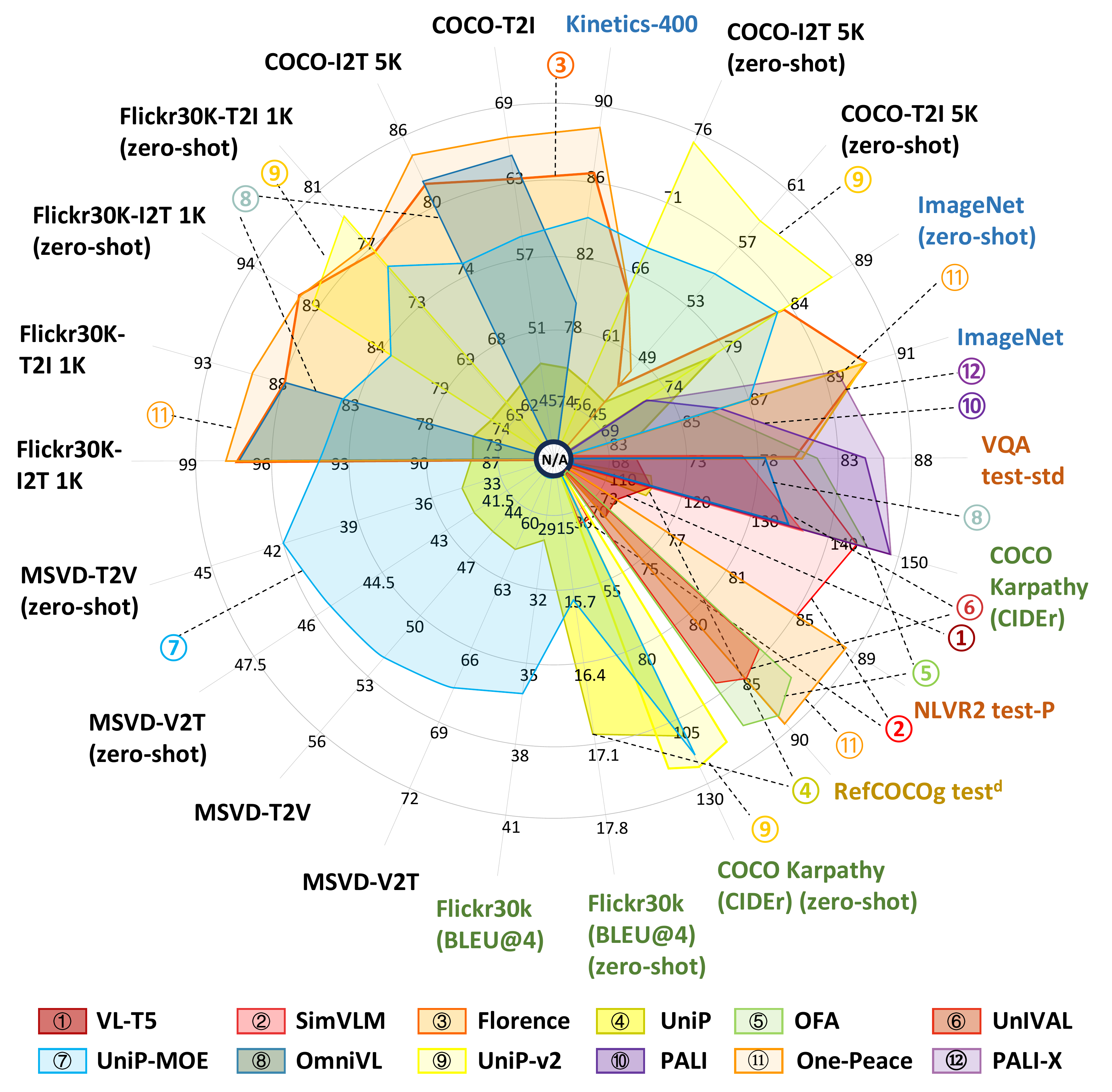}
	\centering
    \vspace{-10pt}
	\caption{Comparisons of the performance for different generalist models \cite{cho2021vlt5, wang2021simvlm, yuan2021florence, zhu2022uniperceiver, wang2022ofa, shukor2023unival, zhu2022uniperceivermoe, wang2022omnivl,li2023uniperceiverv2,chen2022pali,wang2023one-peace,chen2023pali-x} on several classical benchmarks.}
	\label{fig:modelComparing}
    \vspace{-10pt}
\end{figure}

\subsection{Representative Results Analysis}
Given the wide range of benchmarks used by existing vision generalist models and the varying experimental settings, a comprehensive comparison across all benchmarks is not feasible. Unifying evaluation metrics in this research field would be valuable for providing a thorough and fair comparison of existing models. In this subsection, we conduct a representative analysis, comparing 10 vision generalist models across 22 benchmarks, as illustrated in Figure~\ref{fig:modelComparing}.

These 22 benchmarks are grouped into five categories, each represented by a distinct color: (1) Multimodal retrieval, which includes image-to-text, text-to-image, video-to-text, and text-to-video retrieval tasks on COCO, Flickr30K, and MSVD datasets; (2) Vision-only perception, comprising image classification on ImageNet and video classification on Kinetics-400; (3) Other vision-language reasoning tasks, such as image captioning on COCO Karpathy and Flickr30K datasets, visual question answering on the VQA dataset, visual reasoning on the NLVR2 dataset, and referring expression comprehension on the RefCOCOg dataset.

In our collected benchmarks, One-Peace and Uni-Perceiver-v2 demonstrate relatively high performance on image-text retrieval tasks, while Uni-Perceiver-MoE excels in video-text retrieval tasks. Florence stands out in vision-only perception tasks, whereas PALI and PALI-X deliver exceptional results on the VQA task. While all models are capable of handling a wide range of benchmarks simultaneously, no single vision generalist model has emerged as the best across all tasks.

\section{Technique}
\label{sec:technique}

A prominent hurdle for universal vision models lies in integrating multiple modal data and tasks within one central model. Specifically, the simultaneous processing of input and output across all modalities and tasks is a requirement. The data must be coalesced into a shared and unified space to enable the model to manage it in a consistent manner. Within this section, we will explore several strategies for addressing the complexities associated with multi-domain input, model design, and multi-task output challenges. It is important to note that some models discussed in this section may overlap with those presented in Section~\ref{sec:framework}. However, the distinction lies in the focus of each section: while Section~\ref{sec:framework} emphasizes a broader architectural overview, this section highlights the specific, pioneering techniques that make these models both innovative and representative of current advancements.

\subsection{Multi-Domain Input}
Firstly, the key concept to grasp is that despite the superficial disparities of various domains (textual, visual, auditory, or other), it is essential and also challenging to translate these into a unified feature representation. This shared feature space serves as the foundation for data processing, analysis, and comprehension executed by vision generalist models.

The diverse modality of inputs may range from static images to dynamic video sequences, and from single RGB images to multimodal combinations. Additionally, the granularity of computer vision tasks varies significantly. For instance, at the image level, tasks include image classification and captioning, while region-level and pixel-level tasks encompass object detection and semantic segmentation, respectively. Therefore, the unified data encoding techniques are also expected to thoroughly exploit knowledge in multimodal data with different granularities.

\subsubsection{Multi-domain Input in Single Modal}

\noindent\textbf{Shared Transformer-based methods.} 
MulT~\cite{bhattacharjee2022mult} encodes input images from multiple domains into a shared representation space using a unified Transformer-based encoder across all tasks. EQ-Paradigm~\cite {yang2022unified} proposes the Embedding-Querying Paradigm for 3D understanding tasks, including detection, segmentation, and classification. EQ-Paradigm uniformly transforms the input point cloud into the same feature space, adding specific query position embeddings to adapt to different task requirements.


\vspace{4pt}
\noindent\textbf{Visual Prompt.} 
Inspired by prompt-based learning in natural language processing (NLP), Visual Prompting~\cite{bar2022visual} extends this concept to the visual domain. It operates by providing a reference input-output image pair for a novel task at test time, along with a new input image, with the objective of automatically generating an output image that aligns with the given example. Visual prompt tackles this problem with a simple image inpainting method, which fills a hole in connected images of visual cues. This worked well as long as the inpainting algorithm was trained on the correct data. By conducting masked image modeling over joint input-output image pairs, Painter~\cite{wang2023images} enables task execution conditioned on partial visual information. This allows the model to infer the task from provided input-output examples during inference.


\subsubsection{Multi-domain Input in Multimodal}

\noindent\textbf{Specific Encode.}
Florence~\cite{yuan2021florence} adopts a dual-tower design comprising separate encoders for visual and textual modalities. The image encoder utilizes hierarchical vision Transformers, while the text encoder is built upon conventional Transformer-based architectures. To train the model, Florence employs Unified Contrastive Learning (UniCL)~\cite{yang2022unified}, which extends standard contrastive learning by allowing multiple images to correspond to the same text, treating all such pairs as positives. Data2vec~\cite{baevski2022data2vec} introduces a modality-agnostic learning paradigm applicable to speech, language, and vision. Instead of predicting local or modality-specific targets, it learns to generate contextualized latent representations that summarize the entire input. This is achieved via a self-distillation framework, where a masked input view is used to predict the full input’s latent representation using a shared Transformer backbone. Each modality has its own input encoding: images are decomposed into patch sequences, audio is processed with a 1D convolutional front-end to generate waveform representations, and text is tokenized into sub-word embeddings. OFASYS~\cite{bai2022ofasys} designs modality-specific encoders and integrates them into a unified architecture via adapter modules. It tokenizes text and discretizes image inputs into code representations. A built-in scheduler dynamically assigns data to appropriate preprocessing modules and encoders based on input type and modality. LaVIT~\cite{jin2023unified} introduces a visual tokenizer that converts images into discrete token sequences of variable lengths. These tokenized representations enable large language models (LLMs) to effectively interpret and reason over visual information, facilitating unified multimodal understanding.

\vspace{4pt}
\noindent\textbf{Unified Encode.} 
UNITER~\cite{chen2020uniter} is a unified image-text representation model trained on four large-scale vision-language datasets. It aligns multimodal inputs in a shared embedding space using four pre-training objectives: masked language modeling, masked region modeling, image-text matching, and word-region alignment. Conditional masking is employed to enhance learning efficiency. In UFC-BERT~\cite{zhang2021ufc}, both control instructions and synthetic visuals are represented as discrete token sequences and fed into a Transformer. Codebook \cite{duan2022multi} proposes a method for aligning signals from different modalities using a learnable codebook. Codebook treats images and text as two "views" of the same entity and encodes them into a joint visual-linguistic encoding space spanned by cluster-centric dictionaries. Unified-IO~\cite{lu2022unified} homogenizes each supported input and output into a sequence of discrete lexical tokens, allowing a common representation for all tasks. Text inputs are tokenized using SentencePiece. Image inputs are encoded as discrete tokens using a VQVAE~\cite{van2017neural}. To encode sparse structures such as bounding boxes and keypoints, 1000 additional tokens are added to the vocabulary to denote quantized image coordinates. OMNIVORE~\cite{girdhar2022omnivore} unifies vision modalities by encoding images, videos, and single-view 3D data into a shared embedding space. It transforms images into patches, videos into spatio-temporal tubes, and 3D views into paired RGB and depth patches, all processed by a Transformer. M2-Mix~\cite{so2022geodesic} improves representation learning by mixing image and text features in a hyperspherical embedding space. This approach generates hard negatives across datasets, enhancing feature alignment, distribution uniformity, and transferability to downstream tasks. Angular distances within the hypersphere guide the model in learning more robust cross-modal representations. GPNA~\cite{rahaman2022general} observes that the main difference between data of different modalities lies in location, time, and spectral information, so a general sequence tokenizer module is designed. Each token is formed by concatenating a Fourier-encoded spatial-temporal location with a normalized multi-spectral image patch. The encoded positions and times provide positional information for each marker in the sequence. Normalized multispectral patches contain information about the spectral bands captured by the observations, allowing efficient handling of diverse data modalities in Geospatial Systems. Uni-Perceivers~\cite{zhu2022uni,li2023uniperceiverv2} unify inputs and outputs from diverse modalities by employing a modality-agnostic Transformer encoder, paired with lightweight, modality-specific tokenizers that project data into a shared representation space. This enables the model to process inputs from different modalities in a unified encoding, enabling efficient collaboration between tasks. Perceiver-IO~\cite{jaegle2021perceiver} maps arbitrary input arrays to arbitrary output arrays in a domain-independent process by using cross-attention multiple times. Most of the computation occurs in the latent space, whose size is usually smaller than the input and output, making the process computationally tractable even for very large inputs and outputs. ONE-PEACE~\cite{wang2023one-peace} employs modality-specific adapters for vision, audio, and language to unify features, followed by a shared self-attention mechanism and distinct feed-forward networks (FFNs) for each modality. This “separated-sharing-separated” design enables flexible branching for multimodal tasks.

\subsection{Model Design}

\subsubsection{Transformers}
The Transformer~\cite{vaswani2017transformer} architecture, originally introduced as a breakthrough in Natural Language Processing (NLP), has transcended its linguistic origins to become a game-changer in the realm of computer vision. The advent of the Transformer architecture marked a paradigm shift in NLP, replacing recurrent and convolutional networks as the backbone of language models. The central innovation of Transformers is the self-attention mechanism, capturing global dependencies among input tokens efficiently.

The success of the Transformer architecture hinges on its innovative design, particularly the self-attention mechanism, which enables it to capture both local and global dependencies within input data efficiently. In this section, we delve deeper into the key components that constitute Transformers and their relevance to computer vision~\cite{dosovitskiy2020image, liu2021swin, liu2022videoswin, touvron2021deit}.

\vspace{4pt}
\noindent\textbf{Self-Attention Mechanism.}
The self-attention mechanism is central to Transformer architectures, enabling the model to dynamically weigh the relevance of different input elements. In computer vision, this mechanism manifests through two key components: 1) Multi-Head Attention, which applies self-attention in parallel across multiple subspaces using learned projections, allowing the model to capture diverse contextual relationships; and 2) Scaled Dot-Product Attention, which computes attention scores via dot products between query and key vectors, scaled by the square root of the key dimension to stabilize gradients during training.

\vspace{4pt}
\noindent\textbf{Positional Encoding.}
Unlike recurrent or convolutional neural networks, Transformers lack an inherent notion of input order. To address this, positional encodings are added to the input embeddings to inject sequence information. In computer vision, positional encodings~\cite{wu2021rpe, chu2021crpe} perceive spatial relationships within images, such as the proximity of pixels or the arrangement of objects.

\vspace{4pt}
\noindent\textbf{Feedforward Networks.}
After the self-attention mechanism, Transformers employ feedforward neural networks for further processing. These feedforward networks consist of fully connected layers and activation functions, such as ReLU~\cite{agarap2018relu}. They play a vital role in enabling Transformers to capture complex, non-linear relationships within the input data.

\vspace{4pt}
\noindent\textbf{Layer Normalization and Residual Connections.}
Layer normalization~\cite{ba2016layernorm} and residual connections~\cite{he2016resnet} are essential for stabilizing training in the Transformers. Layer normalization is applied before each sub-layer (including self-attention and feedforward layers), normalizing the activations. Residual connections allow gradients to flow more effectively during training, addressing the vanishing gradient problem.

\vspace{4pt}
\noindent\textbf{Sequence-to-Sequence Architecture.}
While Transformers were originally designed for sequence-to-sequence\\~\cite{sutskever2014seq2seq} tasks in NLP, their sequence-to-sequence architecture has found relevance in computer vision. It enables end-to-end processing of sequences, excelling at tasks like image captioning or video analysis, where input data is inherently sequential.

An intriguing characteristic of Transformer models is their versatility in handling a wide spectrum of sequence-to-sequence tasks. This adaptability stems from the fundamental architecture of Transformers, which relies on self-attention mechanisms to process sequential data. In essence, most tasks, regardless of their complexity or domain, can be transformed into a sequence-to-sequence format. By using dedicated sub-modules to convert diverse data types into a uniform sequence structure, the Transformer framework gains a distinct advantage in tackling these tasks. This inherent modularity allows for the seamless integration of specialized components, making it exceptionally well-suited for various applications.

\subsubsection{Diffusion Models}
Diffusion models~\cite{ho2020denoising}, once relegated to the periphery of machine learning, have recently gained prominence as a versatile and powerful class of generative models. Originally introduced as a novel framework for denoising, they have since evolved into a transformative paradigm with applications spanning image generation, data completion, and more. Diffusion models originated as a solution to the problem of denoising data. Over the years, they have undergone significant refinement~\cite{rombach2022ldm, nichol2021ddim, lu2022dpmsolver, nichol2021glide, ho2022classifierfree}, transitioning from their rudimentary forms to the sophisticated models we have today. Central to this evolution is the notion of diffusion processes, wherein noise is gradually added to an image or data point, and the model learns to iteratively recover the original, noise-free data. This elegant idea has catalyzed the development of state-of-the-art generative models that are both effective and efficient.

At the heart of diffusion models lies the concept of probabilistic modeling via diffusion processes. The core principles encompass the following key elements: \textbf{1) Diffusion Process}: The gradual introduction of noise into data is a fundamental concept. Diffusion models leverage this process to learn a powerful likelihood function. \textbf{2) Inference Networks}: Diffusion models often employ neural networks to perform inference, enabling them to estimate the noise levels and generate high-quality samples. \textbf{3) Sampling}: Diffusion models excel in generating samples from complex data distributions. By iterative denoising, they produce realistic and high-fidelity samples.

The versatility of diffusion models extends across a spectrum of machine learning tasks. They have made significant contributions in various domains, including: \textbf{1) Image Generation}~\cite{rombach2022ldm, zhang2023controlnet}: Diffusion models have emerged as formidable contenders in the field of image generation, capable of creating photorealistic images from limited data. \textbf{2) Data Imputation}~\cite{zheng2022diffimputation, kotelnikov2023tabddpm}: They excel in data imputation tasks, where missing or corrupted data must be restored with high accuracy. \textbf{3) Anomaly Detection}~\cite{wolleb2022diffusionad, wyatt2022anoddpm}: Diffusion models are increasingly being leveraged for anomaly detection, detecting outliers or irregularities in data. \textbf{4) Semi-Supervised Learning}~\cite{you2024diffusionssl}: They have found applications in semi-supervised learning, enabling models to learn from both labeled and unlabeled data efficiently. \textbf{5) Representation Learning}~\cite{zhong2023learning}: Diffusion models facilitate effective feature learning, capturing meaningful latent features from complex data distributions.

Diffusion models use probabilistic modeling to assess how likely data is under different levels of noise. This probabilistic approach grasps intricate data patterns and empowers diffusion models to uncover insights across diverse datasets and tasks, showcasing their remarkable potential in cross-modal assignments. 

\subsubsection{Mixture of Experts (MoE)}

In recent years, increasingly complex tasks in computer vision demand the modeling of intricate data distributions and the capture of fine-grained patterns within visual data. Traditional neural networks, while powerful, often struggle to handle these complexities effectively. This sets the stage for the emergence of the MoE~\cite{gormley2019mixture} framework as a compelling paradigm in computer vision.
The motivation behind employing MoE in computer vision stems from the need to address these complexities. MoE models offer a solution by breaking down tasks into subtasks, each managed by an expert network. This paradigm leverages the collective wisdom of experts to make informed predictions, making it promising for tackling various challenges simultaneously.

At the core of the Mixture of Experts (MoE) framework lies a well-defined mathematical formulation. Let's denote our input data as $x$ and the expert networks as $E_1, E_2, \ldots, E_k$, where $k$ represents the number of experts. The primary idea is to create a weighted mixture of the expert networks' predictions based on input data. This can be represented as $M_i(x) = E_i(x)$, where $M_i(x)$ represents the output of the $i$-th expert network for the input $x$. These outputs are combined to form the final prediction $P(x)$ through weighted summation:

\begin{equation}
    P(x) = \sum_{i=1}^{k} \alpha_i(x) \cdot M_i(x)
\end{equation}

The mixture weights $\alpha_i(x)$ are determined by mixture gates or switches, which can be modeled as a softmax operation over gating logits:

\begin{equation}
    \alpha_i(x) = \frac{e^{g_i(x)}}{\sum_{j=1}^{k} e^{g_j(x)}}
\end{equation}

The gating logits $g_i(x)$ are typically learned from the input data and control the contribution of each expert to the final prediction. Training the MoE model involves optimizing the parameters of the expert networks and the gating mechanism to minimize a suitable loss function for the specific task at hand. This formulation elegantly leverages a mixture of specialized experts to make predictions based on input data.

The inherent characteristics of Mixture of Experts (MoE) models have positioned them as a formidable paradigm, particularly well-suited for multi-task learning scenarios. By modeling multiple expert branches, MoE models exhibit the unique ability to specialize in distinct tasks. This specialization not only enables significant parameter and computational savings but also introduces a more principled approach to modeling. The MoE framework inherently acknowledges that different tasks may require specialized expertise, and this recognition aligns with the practical demands of multi-task learning.
In contrast to traditional single-task models, MoE's design stands as a distinct advantage in the realm of general-purpose multi-task models. The specialized expert branches not only enhance task-specific performance but also facilitate efficient knowledge transfer across tasks. This inherent modularity allows for the incorporation of domain-specific knowledge or architectural variations within each expert, further enhancing the adaptability of MoE models.
Moreover, MoE models are inherently equipped to handle tasks with varying complexities and data distributions. This flexibility is particularly valuable in applications where tasks span a spectrum of domains, from natural language processing to computer vision and beyond. The efficiency gains achieved through parameter sharing and the principled approach to expertise specialization underscore the transformative potential of MoE models in multi-task learning scenarios.

In summary, the unique characteristics of MoE models~\cite{fedus2022switchtransformer, lewis2021base, du2022glam, riquelme2021scalingmoe}, rooted in the modeling of multiple expert branches, empower them to excel in multi-task learning settings. Their ability to efficiently allocate resources, specialize in tasks and adapt to diverse domains positions MoE models as a compelling choice for the development of general-purpose multi-task models, offering both practical advantages and sound modeling principles.

\subsection{Multi-Task Output}
The output of vision tasks also has different formats. Such responses might be spatial data, exemplified by edges, bounding boxes, and masks, or they might represent semantic content, such as categorizing labels or descriptions. 

At the beginning, only specific models perform specific tasks, so the easiest way is to directly train multiple models together to obtain the requirement that one model can complete multiple tasks. But the amount of calculation and parameters of such models are very large, and the training cost is unacceptable. Later, researchers found that they could design a model with a stronger expressive ability to learn multiple tasks, share the same model, and only use specific heads to decode different tasks. 
In the preceding section, we will delve into the specifics of how these universal features can be transmuted into formats that align with the requirements of varying tasks.

\subsubsection{Specific Decode}
Mask R-CNN~\cite {he2017mask} is a pioneering unified model that realizes multi-tasking in vision, including instance segmentation and object detection. Mask R-CNN first performs object detection to obtain the corresponding anchor box, and then performs binary classification on the features in the box to obtain the corresponding mask. Mask DINO~\cite{li2023mask} designs different decoders for semantic segmentation, instance segmentation and object detection for one shared model. To support multiple tasks, AiT~\cite{ning2023all} uses separate VQ-VAE models for depth and instance segmentation, predicting token sequences conditioned on the input task token. For different tasks, MultiMAE~\cite{bachmann2022multimae} uses separate decoders for each task in order to reconstruct masked markers from visible markers. MultiMAE supports tasks such as image classification, semantic segmentation, and depth estimation. For each task, a corresponding decoder is required, and the computational cost of the model scales linearly with the number of tasks. MulT~\cite{bhattacharjee2022mult} employs task-specific decoders with identical architectures but distinct parameter sets, appending task-specific heads to each decoder. For example, a model trained for both semantic segmentation and depth prediction would have two heads: one for semantic segmentation with $K$ channels and a softmax, and another for depth estimation with a single channel and a sigmoid.

MetaTransformer~\cite{zhang2023metatransformer} learns representations from the input data and then feeds the representations to the task-specific heads, which mainly consist of multi-layer perceptrons (MLPs). Its task-specific heads vary with the modalities and tasks involved. OFASYS \cite{bai2022ofasys} designs different decoding methods for different modalities and adds all decoders as adapters to the general network, which is generally the reverse process of encoders, such as converting tokens into text, converting code into image, etc. OFASYS  system includes a scheduler that automatically routes the data in each slot to the appropriate postprocessors and adapters based on slot type or mode. UniT~\cite{hu2021UniT} applies task-specific prediction heads to different decoder states for each task and divides the tasks into three kinds of heads to achieve: class head, box head, and attribute head. 12 in 1~\cite{lu202012} groups a variety of tasks (vocab-based VQA, image retrieval, referring expressions, and multimodal verification), with each task having a task-specific head. The aligned image-caption pair scores $h_{IMG}$ and $h_{CLS}$ can be obtained through the shared model:
\begin{equation}
    Rel(I,Q) = W_i(h_{IMG}\odot h_{CLS})
\end{equation}
where $W_i\in \mathbb{R} ^{d\times 1}$ is shared across COCO and Flickr30k image retrieval tasks.

\newcommand{\paragrapha}[2][3pt]{\vspace{#1}\noindent\textbf{#2}}

\begin{table}[!t]
\center
\caption{\textbf{Framework and technique of vision generalist model.} We report the framework and technical categories of each VGM work. In the technology category, we consider the diversity of input modalities, the number of downstream tasks, and the design of the model architecture. For input modalities, where "T" represents Text, "I" represents Image, "V" represents Video, "A" represents Audio, "S" represents sparse data, such as boxes, keypoints, etc., "ST" represents structured data, such as tables, etc., "P" represents point clouds, and finally "O" represents data modalities that are not commonly used in visual models, such as time series and actions. For the column "Head", we also list whether task-specific heads are needed or not.
}
\vspace{-10pt}
\label{tab:technique_conclusion}
\setlength\tabcolsep{1pt}
\resizebox{1\linewidth}{!}{
\begin{tabular}{ccccccc}
\toprule
Method&Publication& Inputs&Task &Model&Head\\ %
\midrule
\multicolumn{6}{c}{\textit{Encoding-based}}\\
\midrule
UniT~\cite{hu2021UniT}&CVPR 2021&T-I&7&Transformer&\checkmark\\
Meta-Transformer~\cite{zhang2023meta}&Arxiv&T-I-A-V&16&Transformer&\checkmark\\
&&S-ST-P-O&&&\\
MultiMAE~\cite{bachmann2022multimae}&ECCV 2022&I&3&Transformer&\checkmark\\
Perceiver~\cite{jaegle2021perceiver}&PMLR 2021&I-V-A-P&5&Transformer&\ding{55}\\
Perceiver-IO~\cite{jaegle2021perceiverio}&ICLR 2022&T-I-V-A&14&Transformer&\ding{55}\\
HiP~\cite{carreira2022hip}&Arvix&T-I-V-A&5&Transformer&\ding{55} \\
Uni-Perceiver~\cite{zhu2022uniperceiver}&CVPR 2022&T-I-V&8&Transformer&\ding{55}\\
Uni-Perceiver v2~\cite{li2023uniperceiverv2}& CVPR 2023&T-I&6&Transformer&\ding{55}\\
Uni-Perceiver-MoE~\cite{zhu2022uniperceivermoe}&NeurIPS 2022&T-I-V&7&MoE&\ding{55}\\
GenLV~\cite{chen2024genlv}&ACM MM 2024&I&10&Transformer&\ding{55}\\
EQ-Paradigm~\cite{yang2022unified}&CVPR2022&P&4&Transformer&\checkmark\\
Pix2Seq V2~\cite{chen2022unified}&NeurIPS 2022&T-I-S&4&Transformer&\ding{55}\\
UViM~\cite{kolesnikov2022uvim}&NeurIPS 2022&I&3&Transformer&\ding{55}\\
MAE-VQGAN~\cite{bar2022visual}&NeurIPS 2022&I&4&Transformer&\ding{55}\\
Painter~\cite{wang2023images}&CVPR 2023&I&5&Transformer&\ding{55}\\
Florence~\cite{yuan2021florence}&Arxiv&T-I-V&7&Transformer&\checkmark\\
ONE-PEACE~\cite{wang2023one-peace}&Arxiv&T-I-A&12&Transformer&\ding{55}\\
OneFormer~\cite{jain2023oneformer}&CVPR 2023&T-I&3&Transformer&\ding{55} \\
\midrule
\multicolumn{6}{c}{\textit{Sequence-to-Sequence}}\\
\midrule
VL-T5~\cite{cho2021vlt5}&ICML 2021&T-I&12&Transformer&\ding{55} \\
SimVLM~\cite{wang2021simvlm}&ICLR 2022&T-I&5&Transformer&\ding{55}\\
OmniVL~\cite{wang2022omnivl}&NeurIPS 2022&T-I-V&8&Transformer&\ding{55}\\
KOSMOS-1~\cite{huang2023kosmos1}&NeurIPS 2023&T-I&10&Transformer&\ding{55}\\
KOSMOS-2~\cite{peng2023kosmos2}&Arxiv&I&7&Transformer&\ding{55}\\
UnIVAL~\cite{shukor2023unival}&TMLR 2023&T-I-V&8&Transformer&\checkmark\\
VisionLLM~\cite{wang2024visionllm} & NeurIPS 2023 & T-I & 4 & Transformer & \checkmark\\
VisionLLMv2~\cite{wu2024visionllmv2} & NeurIPS 2024 & T-I-V & 7 & Transformer & \ding{55} \\
GiT~\cite{wang2024git}&ECCV 2024&T-I-S&5&Tranformer&\ding{55} \\
Chameleon\cite{team2024chameleon}&Arxiv&T-I&5&Transformer&\checkmark\\
MiniGPTv2~\cite{chen2023minigptv2} & Arxiv & T-I-V & 9 & Transformer & \ding{55} \\
LAVENDER~\cite{li2023lavender}&CVPR 2023&V-I&4&Transformer&\ding{55}\\
OFA~\cite{wang2022ofa}&ICML 2022&T-I-S&8&Transformer&\ding{55}\\
OFASys~\cite{bai2022ofasys}&Arxiv&T-I-A-V&23&MoE&\ding{55}\\
&&M-S-ST&&&\\
Unified-IO~\cite{lu2022unified}&ICLR 2023&T-I-S &22&Transformer& \ding{55}\\
Unified-IO-2~\cite{lu2024unifiedio2} & CVPR 2024 & T-I-V-A & 35 & Transformer & \ding{55} \\
PixelLM~\cite{ren2024pixellm} & CVPR 2024 & T-I & 7 & Transformer & \ding{55}\\
PALI~\cite{chen2022pali}&ICLR 2023&T-I&6&Transformer&\checkmark\\
PaLI-X~\cite{chen2023palix}&CVPR 2024&T-I&6&Transformer&\checkmark\\
LVM~\cite{bai2024lvm} & CVPR 2024 & T-I-V & 7 & Transformer & \ding{55}\\
Flamingo~\cite{alayrac2022flamingo}&NeurIPS 2022&T-I-V&16&Transformer&\ding{55}\\
InternGPT~\cite{liu2023interngpt}&Arxiv&T-I-A-V-S-O&--&Transformer&\checkmark\\
GATO~\cite{reed2022gato}&TMLR&T-I-V-O&12&Transformer&\ding{55} \\
PaLM-E~\cite{driess2023palme}&Arxiv&T-I-O&14+&Transformer&\ding{55} \\
RT-2~\cite{brohan2023rt}&Arxiv&T-I&--&Transformer&\ding{55}\\
EmbodiedGPT~\cite{mu2024embodiedgpt}&Arxiv&T-I&5&Transformer&\ding{55}\\
LEO~\cite{huang2023leo}&ICML 2024&T-I-P&8&Transformer&\ding{55}\\
\hline
\end{tabular}
}
\vspace{-10pt}
\end{table}

\subsubsection{Unified Decode}
Maskformer~\cite{cheng2021per} and Mask2former~\cite{cheng2022masked} use the general Transformer framework to unify the semantic segmentation and instance segmentation tasks into a task query to obtain different mask expressions. OneFormer~\cite{jain2023oneformer} introduces a unified Transformer-based framework to handle diverse image segmentation tasks. The model employs task tokens to specify different tasks and facilitate dynamic task inference. Additionally, it incorporates task-conditioned query formulation combined with a query-text contrastive loss, enabling more precise inter-task and inter-class distinctions. Pix2Seq-D~\cite{chen2022generalist} converts the panoptic segmentation of images and videos into mask generation tasks and uses the diffusion model to implement this generation process. Pix2seq v2~\cite{chen2022unified} uses prompt learning to specify tasks by feeding them into the network. By standardizing the output format for each task, the sequence output adapts to the prompt, enabling the generation of task-specific results based on the provided description. OFA~\cite{wang2022ofa}, inspired by Pix2Seq v2, discretizes text, images, and objects into tokens using a unified vocabulary that includes components like quantized image tokens and sparse visual representations. UNINEXT~\cite{yan2023uninext} consolidates multiple instance perception tasks into a unified object discovery and retrieval framework, with each task being differentiated by specific prompts. The process of obtaining results is framed as a retrieval task based on the proposed instance-prompt matching scores. OMG-Seg~\cite{li2024omg} further extends this unification by encompassing a broad spectrum of segmentation tasks, addressing various visual modalities, different perception granularities, and challenging interactive or open-vocabulary scenarios. All task outputs are unified into a single query representation, and a shared Transformer decoder is employed to exploit the relationships between task queries and visual features.

Uni-Perceivers~\cite{zhu2022uni,li2023uniperceiverv2} formulates different tasks as a unified maximum likelihood estimation problem. Given an input $x\in \mathcal X$ and the candidate target set $\mathcal Y$, the objective is to identify $\hat y \in \mathcal Y$ with the maximum likelihood as
\begin{equation}
    \hat{y} = arg\max_{y\in \mathcal{Y}} P(x,y)
\end{equation}
where $P(x,y)$ is the joint probability distribution. The joint probability is the cosine similarity between the representations of $x$ and $y$:
\begin{equation}
    P(x,y) \propto \exp(\cos(f(x),f (y))/\tau) 
\end{equation}
where $f (\cdot)$ is the Transformer encoder, and $\tau > 0$ is a learnable temperature parameter. For outputs with a multi-task or multimodal structure, Perceiver-IO~\cite{jaegle2021perceiver} learns a query for each task or modality. This information allows the network to distinguish one task or modality query from others, just like positional encoding allows attention to distinguish one location from another. Unified-IO~\cite{lu2022unified} converts features into discrete token sequences and runs different decoding modules according to distinct task representations, such as the VQVAE decoder for image segmentation, and the SentPiece decoder for VQA or object detection.

Recently, InstructDiffusion~\cite{geng2024instructdiffusion} presents a novel generalist modeling interface that integrates various perception and generation tasks into an intuitive image manipulation process. The model leverages the full potential of the Diffusion~\cite{rombach2022ldm} framework, directly generating results in the continuous pixel space, conditioned on specially designed task instructions. This work paves the way for future research on vision generalist models by framing generalization as a unified task.

\section{Related Domains}
\label{sec:related-domains}

In this section, we present an exposition of the three most pertinent sub-fields within the context of the vision generalist model: multi-task learning, vision-language learning, and open vocabulary learning. They boast extensive developmental histories and can be perceived as foundational steps of the vision generalist model.

\subsection{Multi-Task Learning}

In the field of machine learning, the pursuit of constructing models capable of efficiently learning from limited data and generalizing across diverse tasks has propelled the development of innovative techniques. One technique that has garnered significant attention is multi-task learning (MTL). Unlike traditional single-task learning approaches that concentrate on solving a singular and specific problem, multi-task learning aims to enhance the learning process by simultaneously considering multiple related tasks. This approach leverages shared information and underlying relationships among tasks to enhance the model's overall performance and generalization capabilities. The motivation behind multi-task learning originates from the recognition that many real-world problems involve interdependencies among multiple tasks. Frequently, these tasks share common underlying features, and addressing them separately may lead to sub-optimal results due to redundant feature extraction and inadequate data utilization. Multi-task learning exploits the shared patterns, enhancing the performance on individual tasks via collective learning.

Architectural design plays a pivotal role in achieving a balance in performance across target tasks. Traditional multi-task learning methods employ task-specific decoders to address distinct tasks~\cite{zhang2014facial,dai2016instance,zhao2018modulation} or incorporate interactive modules between each task~\cite {misra2016cross,ruder2019latent}. The multimodality encoder functions as a unifying mechanism across datasets or data modalities, enabling the core architecture to handle heterogeneous data and produce responses for multiple tasks~\cite{nguyen2019multi}. ~\cite{liang2018evolutionary,gao2020mtl,chen2021autoformer} design learnable architectures, for instance, neural architecture search methods to establish more effective connections for multi-task learning, guided by the overall model performance. Another innovative approach, as employed by~\cite{bengio2013estimating,andreas2016neural,rosenbaum2017routing}, involves utilizing a dynamic architecture that bridges the gap between inputs and tasks. Optimization in multi-task learning provides another perspective for managing and harmonizing the influence among target tasks. In early multi-task optimization endeavors, adjustments and scaling were applied to the output loss functions of different tasks. This involved amalgamating a series of partial loss functions into a comprehensive overall loss function within the model~\cite{kendall2018multi,chen2018gradnorm,liu2019end,guo2018dynamic}. The notion of soft parameter sharing, as employed by~\cite{duong2015low,yang2016trace}, proposes regularization among parameters. This strategy encourages similarity among parameters across different tasks and proves beneficial in enhancing performance, particularly with multiple small datasets.~\cite{sharma2017learning} introduces the concept of task scheduling. This technique entails learning to select the task to be trained at each step of the training process. Knowledge distillation~\cite{hinton2015distilling} finds utility in various domains such as model compression, pruning, and transfer learning~\cite{polino2018model,rao2023dynamic}. Following this, the multi-task model can assume the role of a student network, guided by multiple teacher networks, each supervised by a distinct task~\cite{rusu2015policy,parisotto2015actor}.  

The fundamental requirement of a vision generalist model is its capacity to manage diverse inputs and outputs across varying domains and formats. Consequently, the advancement of multi-task learning stands as a crucial stride towards realizing a more comprehensive generalist model, which is endowed with greater proficiency across AI tasks and a closer approximation to human behavior.

\subsection{Vision-Language Learning}

In recent times, neural networks have witnessed rapid growth and remarkable achievements in the two primary domains of artificial intelligence: computer vision (CV) and natural language processing (NLP). Central to these advancements lies the exploration of connections between vision and language. Multimodality models emerge as a solution to several practical challenges posed by real-world scenarios. Moreover, they exhibit the potential to surpass single-modality models in terms of representation performance.

Early vision-language learning relies on a spectrum of task-specific challenges, encompassing visual question answering, image captioning, image-text matching, and visual reasoning. In the context of Visual Question Answering (VQA), the objective is to answer questions posed about a given image. Traditional VQA approaches, such as Vanilla VQA~\cite{antol2015vqa}, employ VGG~\cite{simonyan2014very} as the image encoder and LSTM~\cite{hochreiter1997long} as the language encoder. Progressing beyond, subsequent methodologies seek enhanced modality fusion by manipulating image features and language tokens~\cite{fukui2016multimodal,shih2016look,yang2016stacked}. Turning to image captioning, the goal is to generate succinct descriptions encapsulating the essence of provided images. Earlier investigations~\cite{kulkarni2013babytalk,karpathy2015deep,xu2015show} harnessed global feature extractors to encapsulate the semantic meaning of images. Advancements in this field~\cite{anderson2018bottom} introduce attention mechanisms to concentrate on informative regions within images, thereby refining the captioning process.

The development of architectural designs in vision and language tasks further promotes the advancement of vision-language learning. Convolutional Neural Networks (CNN)\cite{krizhevsky2012imagenet,he2016deep} and Recurrent Neural Networks (RNN)\cite{rumelhart1985learning,hochreiter1997long} are widely employed in various vision and language tasks, followed by the Transformer~\cite{vaswani2017attention} architecture, which aims to better utilize the interactions among language tokens. The Vision Transformer~\cite{dosovitskiy2020image} treats images as $16\times 16$ vision tokens and unifies the architectural design in both the vision and language domains. This progress ushers in a new era in vision-language pre-training, achieved by jointly training vision and language models on image-text paired datasets. One fundamental approach to vision-language pre-training is BERT~\cite{devlin2018bert}, which masks language tokens and pre-trains the Transformer encoders to possess a robust representation ability for text embeddings. This high-quality text embedding enables models from other modalities to seamlessly integrate into the language's latent space and achieve cross-modality representation. Numerous pre-training methods~\cite{sun2019videobert,li2019visualbert,chen2019uniter,zhou2020unified} have been proposed based on masked language modeling or image-text matching tasks. They are usually pre-trained on image-text joint datasets such as CC3M~\cite{sharma2018conceptual,changpinyo2021conceptual} and Visual Genome~\cite{krishna2017visual}. 

A significant milestone in the vision-language pre-training domain is CLIP~\cite{radford2021learning}. CLIP has set the trend for models trained with large-scale datasets, resulting in more potent representation capabilities. To train large vision-language models with over 400M image-text pairs, CLIP introduces a novel training strategy that involves jointly learning image and language knowledge. The primary goal of CLIP pre-training is to maximize the similarity between positive pairs while minimizing the impact of negative pairs. Through extensive image-text pair pre-training, CLIP demonstrates superior performance compared to most previous baselines in zero-shot matching scenarios. The pre-trained features and model weights of CLIP serve as robust image and text representations for a wide array of vision and language downstream tasks~\cite{gu2021zero,rao2022denseclip}. In addition to visual and language recognition, researchers have discovered the potential of utilizing weakly-labeled image-text pairs for text-to-image generation tasks. DALL-E~\cite{ramesh2021zero} encodes training images using a discrete auto-encoder and processes image and text data in a joint manner. During the inference phase, the model generates image tokens in an auto-regressive fashion. Subsequent studies~\cite{rombach2022high,zhao2023unleashing} have attempted to model the generation process using a diffusion-based approach. The emergence of extensive image-text corpora like LAION~\cite{schuhmann2022laion} contributes to enhanced generalizability in text-to-image generation.

With the rapid advancements in Large Language Models (LLMs), recent vision-language learning methods have increasingly integrated LLMs to design multimodal generalist models. However, as these methods primarily produce language-based outputs, we exclude them from the scope of generalist vision models, which emphasize the handling of vision-specific tasks. Several prior works concentrate on architectural innovations. Frozen~\cite{tsimpoukelli2021frozen} was among the first to employ a pre-trained, frozen language model for multimodal few-shot learning by introducing a vision encoder that encodes images into sequential prefix prompts. Building on this approach, GIT~\cite{wang2022git} incorporates temporal embeddings to extend the paradigm to video data. LLaMa-adapter~\cite{zhang2024llamaadapter, gao2023llamaadapterv2} introduces learnable adaptation prompts and a zero-initialized attention mechanism, enabling the LLM to adapt quickly to vision-language tasks. To bridge the gap between visual and language representations, BLIP-2~\cite{li2023blip2} pretrains a compact query transformer connecting a frozen image encoder with a large language model. OneLLM~\cite{han2024onellm} proposes a general projection framework featuring modality-specific experts and a routing mechanism for aligning multimodal inputs with language. MiniGPT-4~\cite{zhu2023minigpt4} achieves similar alignment by employing a simple projection layer, enabling effective integration of a frozen vision encoder with GPT-4~\cite{achiam2023gpt4}. LLaVa~\cite{liu2024llava, liu2024llava15} adopts visual instruction tuning to create a versatile visual assistant, reformatting image-text pairs into an instruction-following framework and constructing a large-scale multimodal model (LMM). Additionally, it explores multi-task training strategies and scaling techniques to establish stronger baselines. Other studies focus on extending LLMs to novel vision-related applications. Shikra~\cite{chen2023shikra} tackles the challenge of spatial referential dialogue by constructing specific instructional data for training a unified model. Qwen-VL~\cite{bai2023qwenvl} enhances grounding and text-reading capabilities by aligning image-caption-box tuples. Qwen2-VL~\cite{wang2024qwen2} further supports dynamic resolution processing and proposes multimodal rotary position embedding to enhance cross-modal alignment. The model family is further scaled up to 72B parameters, enabling improved performance across diverse vision-language tasks. Qwen2.5-VL~\cite{bai2025qwen25} achieves substantial gains in visual grounding and long-video understanding, supported by two key innovations: native dynamic resolution perception, which allows scale-aware spatial encoding, and absolute time encoding, which enables precise temporal reasoning over extended video sequences. In addition, a subset of vision-language models has been developed to integrate large language models (LLMs) with video perception and 3D data understanding. For video tasks, models such as Video-LLaVA~\cite{lin2023videollava}, Video-ChatGPT~\cite{maaz2023videochatgpt}, MiniGPT4-Video~\cite{ataallah2024minigpt4video}, and VideoCoCa~\cite{yan2022videococa} couple LLMs with video encoders and temporal projection modules to enable multi-turn dialogue and video-language alignment. For 3D understanding, approaches like PointLLM~\cite{xu2024pointllm} and ShapeLLM~\cite{qi2024shapellm} equip LLMs with point cloud encoders, thereby enabling text-grounded reasoning over 3D spatial data.

\subsection{Open Vocabulary}

Traditional methods' training and evaluation paradigms rely on a closed vocabulary consisting of predefined categories, which limits extensibility. For instance, datasets like COCO~\cite{lin2014microsoft} come with 80 predefined categories, which can be impractical in real-world applications. However, the number of categories in real-world scenarios is extensive and constantly evolving. Consequently, researchers have introduced the concept of open vocabulary learning to tackle this challenge by enabling the recognition of categories beyond the scope of pre-defined labels. Conventional closed vocabulary approaches encounter difficulties when accommodating new or novel objects, leading to suboptimal performance in dynamic environments. Open vocabulary strategies empower models to adapt and identify emerging object categories without the need for frequent updates to a predefined vocabulary.

Open vocabulary learning, initially introduced in the object detection field, represents a core and widely recognized task within computer vision. Mainstream approaches to open vocabulary learning extend their knowledge base through the incorporation of external training data and the utilization of multimodal pre-trained models. The pioneering work OVR-CNN~\cite{zareian2021open} emerged as the first in the field of open vocabulary object detection, harnessing a vast collection of image-caption pairs to enhance performance on previously unknown categories. ViLD~\cite{gu2021zero} distills the insights of the image encoder from multimodal pre-trained models, such as CLIP, into the detector. Detic~\cite{zhou2022detecting} directly trains the detection model using large-scale classification datasets and detection datasets, utilizing a classifier to manage classification data and transfer classification knowledge to detection models. GLIP~\cite{li2022grounded} takes a holistic approach by unifying open vocabulary tasks and visual grounding tasks.  This design significantly bolsters performance in open vocabulary detection and zero-shot detection scenarios. RegionCLIP~\cite{zhong2022regionclip} extends the capabilities of CLIP to region-level data through pre-training and fine-tuning on image regions. This extension enhances CLIP's ability to handle and understand localized information.

The concept of open vocabulary can be extended to segmentation tasks, driven by a similar motivation as observed in object detection. Capitalizing on the rich semantic knowledge present in vision-language pre-trained models, researchers~\cite{ding2022decoupling,zhou2022extract} leverage these models to substitute the original closed-set classifier with textual features. In another approach~\cite{ghiasi2022scaling,liang2023open}, open vocabulary knowledge is extracted from image captions, enabling the model to learn from weakly labeled textual information associated with images. \\Group-ViT~\cite{xu2022groupvit} introduces a novel approach to segmentation. Built on the foundation of the group mechanism, this method employs the image-title contrast loss function during training. The realm of open vocabulary learning extends its potential to diverse tasks, like 3D understanding. In this context, 3Detic~\cite{lu2022open} integrates open vocabulary principles into 3D point cloud recognition. This is achieved by training 3D and 2D detectors on annotated 3D point cloud data, 2D projections of 3D data, and extensive image data.

\section{Conclusions}
\label{sec:conclusions}

In this survey, we review the recent progress of vision generalist models (VGM), introduce the corresponding background, and classify VGM in terms of framework and technology. In addition, we briefly introduce three related areas to promote the development of VGM. However, its development is still in the preliminary exploration stage, despite emerging deployments in real-world scenarios. In this section, we first examine current application domains of VGMs, followed by a discussion of key technical challenges, and conclude with perspectives on future directions.

\subsection{Real-World Applications}

\noindent\textbf{Autonomous Driving.} VGMs are increasingly adopted in autonomous driving systems, where they unify multiple perception and decision-making tasks such as object detection, semantic segmentation, lane boundary estimation, and trajectory prediction. Compared to pipelines composed of discrete task-specific models, VGMs offer improved consistency in perception outputs and more holistic policy learning. For instance, Tesla has developed an end-to-end highway stack, supporting lane changing, parking lot detection, and speed profiling. This system is powered by a vision generalist model that ingests multi-camera video inputs and serves as the core component of Tesla’s Full Self-Driving (FSD) platform.

\vspace{4pt}
\noindent\textbf{Robotics.} As discussed in Section~\ref{sec:system_design}, VGMs are also central to the advancement of embodied AI, enabling robots to perform complex tasks in dynamic environments. A representative example is RT-2~\cite{brohan2023rt}, developed by Google DeepMind, which integrates vision-language-action learning to allow real-world robots to execute high-level natural language commands and carry out diverse manipulation tasks.

\vspace{4pt}
\noindent\textbf{Augmented Reality.} In the context of egocentric vision, VGMs contribute significantly to intelligent augmented reality (AR) systems. Their ability to understand first-person visual scenes and generate interactive, context-aware outputs makes them highly suitable for wearable devices. For example, Meta’s AR glasses project (e.g., Orion) leverages egocentric video input to recognize pantry items and provide personalized cooking recommendations. This functionality is powered by a multimodal, user-adaptive VGM that processes visual and language inputs to support real-time assistance in daily activities.

\subsection{Challenges}

\noindent\textbf{General-Purpose I/O Format.}  
A versatile I/O format holds the potential to streamline various aspects of model design, training methods, and task prompt formulation within the context of general-purpose applications.  In the case of LLM, the prevalent I/O structure adheres to a sequence-to-sequence paradigm.  This design choice allows for the development of model architectures and task prompts that are inherently suited for sequential data.
However, the field of VGM is currently in an early developmental phase, leading to a diversity of I/O formats across different methodologies.  Some approaches in this domain define their I/O format either as sequence-to-sequence, drawing inspiration from NLP, or as image-to-sequence.  
Contrarily, a contrasting viewpoint, championed by Painter~\cite{wang2023images} and MAE-VQGAN~\cite{bar2022visual}, posits that the reliance on sequences might not holistically capture the essence of visual tasks.  Instead, this perspective advocates for an image-to-image I/O format, where tasks are uniformly conceptualized as transformations applied to images.  This standardized image-centric approach demonstrates adaptability across diverse tasks through the consistent utilization of image-based prompts.  Furthermore, certain strategies emerge wherein images are encoded as queries, offering a resolution to the coarse localization challenge prevalent in the aforementioned formats.
Despite these advancements, the field of VGM has yet to converge on a universally applicable I/O format.  As a result, the crucial challenge persists in determining an efficient and all-encompassing Input/Output framework for visual information processing.  This substantial hurdle necessitates continued research and innovation in order to establish a cohesive and broadly applicable solution.

\vspace{4pt}
\noindent\textbf{Data.} 
The training regimen for comprehensive models often necessitates a substantial volume of data, thereby underscoring the pivotal role that both data quality and quantity play in influencing the overall performance of these models.  However, a notable distinction arises when juxtaposing this scenario with that of LLM, given the relatively greater challenge associated with procuring image data as opposed to textual data.
Several factors contribute to this discrepancy.  Primarily, the volume of readily accessible image data available on the Internet pales in comparison to the extensive reservoir of textual data.  This intrinsic imbalance immediately underscores the scarcity of image data in the broader landscape.
Moreover, the inherent nature of textual data lends itself to natural labeling, a facet that significantly streamlines its utility in training processes.  In stark contrast, harnessing image data invariably involves the meticulous human annotation of diverse labels tailored to the distinctive requirements of various vision-related tasks.  This requisite labeling process introduces a layer of complexity that complicates the effective utilization of image data, rendering its integration into model training a considerably more intricate endeavor.
Lastly, the incorporation of image data into model training introduces a host of concerns, encompassing matters of privacy and copyright.  Recent attention to these concerns, notably within the ambit of LLM applications, amplifies the already elevated costs and challenges associated with acquiring image data.  This confluence of privacy and copyright casts a palpable shadow over the procurement process for image data, further underscoring the substantial hurdles encountered when employing such data.

\vspace{4pt}
\noindent\textbf{Generalization and Robustness.} 
Owing to the myriad approaches employed in acquiring image data, the efficacy of vision models often exhibits considerable variance across distinct datasets, owing to the impact of diverse factors such as lighting conditions, weather variations, and camera parameters.  As a consequence, the pursuit of robustness and generalization has emerged as a focal point within the realm of vision studies.
Presently, extant research within the domain of VGM predominantly accentuates the augmentation of model generalization performance across diverse tasks through the formulation of comprehensive task representations.  Notably, a conspicuous gap exists wherein limited attention is directed towards the broader challenge of enhancing the generalizability of these models across a more expansive spectrum of datasets.
Addressing this void demands a concerted effort to delve into the intricacies of cross-dataset generalization, an avenue thus far relatively unexplored.  By bridging this gap, researchers can unravel the nuanced mechanisms that underpin the versatile transference of general models onto a broader array of datasets.

\subsection{Future Directions}
To further advance the development of VGM, we propose here several potential future research directions.

\vspace{4pt}
\noindent\textbf{Efficiency.} The objective of this research trajectory is to cultivate rapid and resource-efficient VGM characterized by minimal operational costs.  This pursuit gains heightened significance in light of the imperatives posed by real-time visual tasks, such as autonomous driving and facial recognition, where expeditious responses are of paramount importance.  The development of agile and efficient VGM thus becomes a compelling aspiration within this context.
Significantly, the deployment landscape for vision models frequently encompasses edge devices, which inherently demand judicious allocation of resources to facilitate seamless operation.  Consequently, the intricate balance between performance and efficiency assumes a central role, precipitating a critical research challenge that merits substantive exploration in the times ahead.

\vspace{4pt}
\noindent\textbf{Evaluation Metric.} 
Contemporary research pertaining to VGM predominantly manifests its efficacy through the exhibition of performance outcomes across diverse downstream tasks.  Nevertheless, a conspicuous void persists in the form of a comprehensive framework for evaluating the multifaceted aptitude of VGM across a spectrum of dimensions, including but not limited to generalization, robustness, reliability, and alignment with societal norms.
The trajectory of future investigations necessitates the creation of an encompassing benchmark alongside a sophisticated set of metrics, which collectively serve as a potent catalyst for propelling the holistic advancement of VGM.  By orchestrating the development of such a comprehensive evaluation apparatus, researchers can systematically delve into the intricacies of VGM performance, thereby fostering a more nuanced comprehension of its multifarious capabilities.

\vspace{4pt}
\noindent\textbf{Learning Paradigm.}
Existing VGMs predominantly prioritize visual perception tasks, such as object recognition and segmentation, while generative tasks—essential for enabling personalization, contextual understanding, and creative applications—remain largely underexplored. Generative tasks hold significant potential for advancing user-centric applications, including content creation, style transfer, and interactive design, yet they are often overshadowed by the focus on perception-based challenges. Among existing efforts, InstructDiffusion~\cite{geng2024instructdiffusion} stands out as a pioneering approach, attempting to reformulate diverse vision tasks within a unified conditional generation framework. This approach not only bridges perception and generation but also underscores the versatility of VGMs in handling complex, context-driven scenarios. Expanding upon this paradigm could foster a more seamless integration of generative capabilities across a broader spectrum of VGM applications, thereby enhancing adaptability and user engagement. Such advancements would further elevate human-computer interaction, enabling the co-creation of novel content while broadening the scope of VGMs in real-world applications.

\vspace{4pt}
\noindent\textbf{Active Learning.}
Future VGMs have the potential to become more proactive by querying users or external databases for additional relevant information when encountering ambiguous tasks or uncertainties. This capability could enable VGMs to dynamically refine their understanding in real-time, leveraging active learning strategies to iteratively improve their performance while reducing dependency on vast pre-existing datasets. Such an approach would not only enhance their efficiency but also allow for more targeted and contextually relevant responses in diverse applications.  A critical challenge in this adaptive process is addressing catastrophic forgetting, which is the tendency of models to lose previously acquired knowledge when learning new tasks or adapting to novel environments. For VGMs to operate effectively in long-term, dynamic real-world scenarios, they must balance adaptability with retention, ensuring that newly acquired skills do not compromise existing capabilities. Developing robust mechanisms for continual learning and knowledge preservation will be essential for equipping VGMs with the flexibility and resilience required to handle the evolving complexities of real-world interactions.

\vspace{4pt}
\noindent\textbf{World Model Knowledge.} Currently, the knowledge embedded in existing VGMs is predominantly acquired through data-driven training processes. While effective, this type of knowledge is often opaque and lacks guarantees of accuracy across diverse scenarios, raising concerns about explainability and reliability. In contrast, real-world applications frequently involve immutable physical laws and domain-specific knowledge that must be consistently upheld. To address this limitation, future VGMs could benefit from integration with world models, frameworks that encapsulate real-world principles and adhere strictly to physical laws. By embedding such structured, rule-based knowledge, VGMs could achieve greater robustness, interpretability, and applicability across a broader range of scenarios. This integration would not only enhance their ability to generalize beyond training data but also ensure their outputs align with fundamental real-world constraints, paving the way for more trustworthy applications in real-life contexts.

\begin{acknowledgements}
This work was supported in part by the National Natural Science Foundation of China under Grant \\ 623B2063, Grant 62125603, Grant 62321005, Grant 62336004, and in part by the National Key Research and Development Program of China under Grant 2022ZD0160102.
\end{acknowledgements}

{\footnotesize
	\bibliographystyle{plainnat}
	\bibliography{egbib}
}

\end{document}